%% file: collas2023_conference.tex
\documentclass{article} 
\usepackage{collas2023_conference,times}
\usepackage{easyReview}

\input{math_commands2.tex}

\usepackage{amssymb}
\usepackage[makeroom]{cancel}
\usepackage{mathletters}
\usepackage{eucal}
\usepackage{minitoc}

\DeclarePairedDelimiter\ceil{\lceil}{\rceil}

\newcommand{\indep}{\perp \!\!\! \perp}
\newcommand{\dfq}{\vcentcolon=}

\usepackage{hyperref}
\hypersetup{
    colorlinks=true,
    linkcolor=red,
    filecolor=magenta,
    urlcolor=blue,
    citecolor=purple,
    pdftitle={Overleaf Example},
    pdfpagemode=FullScreen,
    }

\title{Restarted Bayesian Online Change-point Detection for Non-Stationary Markov Decision Processes}

\author{Reda Alami$^*$ \\
Technology Innovation Institute \\
Abu Dhabi, United Arab Emirates \\
\texttt{reda.alami@tii.ae} \\
\And
\hspace{-20em}Mohammed Mahfoud$^*$\\
\hspace{-20em}Technical University of Munich \\
\hspace{-20em}Munich, Germany\\
\hspace{-20em}\texttt{mo.mahfoud@tum.de} \\
\AND 
\hspace{0em}Eric Moulines \\
\hspace{0em}Ecole Polytechnique \\
\hspace{0em}Paris, France \\
\hspace{0em}\texttt{eric.moulines@polytechnique.edu} \\
}

\preprintcopy 

\begin{document}

\doparttoc 
\faketableofcontents

\maketitle
\def\thefootnote{*}\footnotetext{Equal Contribution.}
\begin{abstract}
We consider the problem of learning in a non-stationary reinforcement learning (RL) environment, where the setting can be fully described by a piecewise stationary discrete-time Markov decision process (MDP). We introduce a variant of the Restarted Bayesian Online Change-Point Detection algorithm (\texttt{R-BOCPD}) that operates on input streams originating from the more general multinomial distribution and provides near-optimal theoretical guarantees in terms of false-alarm rate and detection delay. Based on this, we propose an improved version of the \texttt{UCRL2} algorithm for MDPs with state transition kernel sampled from a multinomial distribution, which we call \texttt{R-BOCPD-UCRL2}. We perform a finite-time performance analysis and show that \texttt{R-BOCPD-UCRL2} enjoys a favorable regret bound of $\order{D O \sqrt{A T K_T \log\left (\frac{T}{\delta} \right)} + \frac{K_T \log \frac{K_T}{\delta}}{\min\limits_\ell \: \infdiv*{\btheta^{(\ell+1)}}{\btheta^{(\ell)}}}}$, where $D$ is the largest MDP diameter from the set of MDPs defining the piecewise stationary MDP setting, $O$ is the finite number of states (constant over all changes), $A$ is the finite number of actions (constant over all changes), $K_T$ is the number of change points up to horizon $T$, and $\btheta^{(\ell)}$ is the transition kernel during the interval $[c_\ell, c_{\ell+1})$, which we assume to be multinomially distributed over the set of states $\sO$. Interestingly, the performance bound does not directly scale with the variation in MDP state transition distributions and rewards, ie. can also model abrupt changes.  In practice, \texttt{R-BOCPD-UCRL2} outperforms the state-of-the-art in a variety of scenarios in synthetic environments. We provide a detailed experimental setup along with a code repository (upon publication) that can be used to easily reproduce our experiments. 



\end{abstract}


\section{Introduction}
In a typical sequential online decision making setting, a decision maker, which we refer to as \emph{agent}, interacts with its environment by first observing its current state. It then select an action from the set of possible actions in its state, moves to another state determined by the stochastic process that generates its state transition distribution, and receives a random reward that quantifies the quality of its current decision/action relative to the set of optimal actions it could have taken in its previous state. Through this continuous interaction, the agent attempts to learn an optimal decision-making scheme or \emph{policy} to maximize its cumulative rewards. This accurately describes the reinforcement learning (RL) problem, which has proven useful in modeling a variety of important problems in many domains.

In classical RL, it is often assumed that state transition distributions and rewards are generated by a stochastic process that is stationary throughout the learning process. However, this assumption is quite restrictive in the context of real online learning environments. Therefore, it is necessary to define a new variant of RL, commonly referred to as non-stationary RL. The study of the latter is supported by a variety of applications in consumer decision modeling (\cite{xu2020reinforcement}), service provider adaptation to customers, and pricing (\cite{taylor2018demand, kanoria2019blind, bimpikis2019spatial, gurvich2019operations}), wireless communication networks (\cite{zhou2015wireless, zhou2016repeated}), epidemic networks and control (\cite{nowzari2016analysis, kiss2017mathematics}), inventory management (\cite{agrawal2019learning, huh2009nonparametric}), to name a few. In the aforementioned application areas, non-stationarity is often due to abrupt changes that can have drastic effects in highly sensitive environments.

While accurately modeling latent change in a non-stationary RL environment is generally very difficult, we are particularly interested in exploring a variant where non-stationarity can be fully modeled by a piecewise stationary discrete-time Markov decision process (MDP). More specifically, we assume a situation in which the associated MDP state transition distributions and rewards can change arbitrarily at unknown predefined time points, which we refer to as \textit{change-points}. In this way, we can accurately and dynamically handle the abruptly changing environment mentioned above. An illustration of the problem can be found in Figure \ref{fig:NSMDP}.
\begin{center}
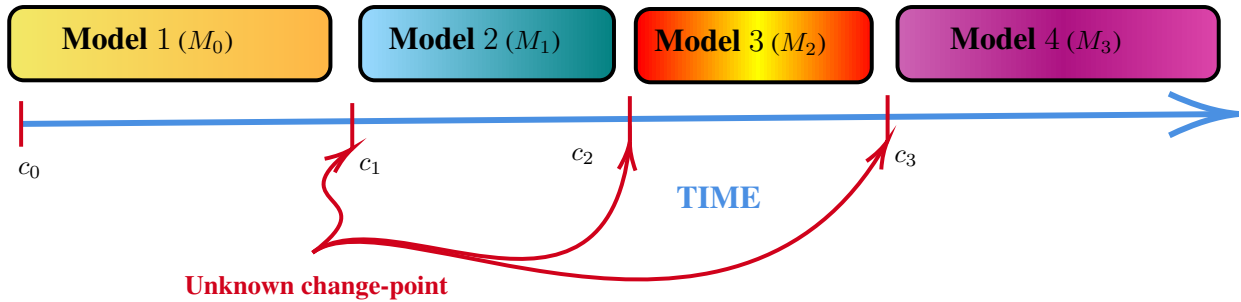
\begin{figure}[H]
    \centering
    \hspace*{-.75cm}
    \scalebox{1}{\input{nonStationaryMDP.tikz}}
    \vspace{-1.7cm}
    \caption{Piecewise Stationary Discrete-time MDP. Starting from change-point $c_\ell$, the environment is modeled by MDP $M_\ell$.}
    \label{fig:NSMDP}
\end{figure}
\end{center}

\vspace{-1.4cm}
\paragraph{Key Contributions:} We list a summary of our main contributions as follows
\begin{enumerate}
    \item We extend the Restarted Bayesian Online Change-Point Detection algorithm \texttt{R-BOCPD} to the more general setting where the online observation stream is generated according to a multinomial distribution, and provide (near) optimal theoretical guarantees in terms of false alarm rate and detection delay control. 
    \item We propose an improved version of \texttt{UCRL2} that incorporates \texttt{R-BOCPD} as a means of detecting changes in the learning environment and provides near-optimal regret bounds that rely only on a small set of past observations and allow for changes of arbitrary magnitude in both state transition distributions and rewards. 
    \item We demonstrate the results experimentally and compare the performance of the algorithm' with that of the state of the art. After publication, a GitHub repository will be provided to ensure reproducibility.
\end{enumerate}


In the following, we list some key results from the literature that relate to our problem.

\section{Related work}
To give a general overview  of past works dealing with a switching RL (MDP RL), we first present some of the main results obtained in the field of stationary vanilla MDP, and then describe some recent key contributions and models in the field of non-stationary RL. Finally, we also present some results (mostly algorithmic in nature) mainly from the time series area.

\vspace{-1em}

\subsection{RL in Stationary MDPs}
We restrict ourselves to contributions that are directly relevant to our problem. In particular, we distinguish the discounted (\cite{sidford2018near, sidford2018variance, wang2020randomized}) and non-discounted reward cases (\cite{auer2008near, azar2017minimax, dann2017unifying, jin2018q, zanette2019tighter}). In the former case, \cite{sidford2018near, sidford2018variance, wang2020randomized} proposed near-optimal algorithms with respect to sample complexity. For the latter case, where the regret of an algorithm $\mathcal{A}$ is defined as the difference between the reward obtained by $\mathcal{A}$ and that of an optimal algorithm, numerous regret bounds have been proposed. \cite{auer2008near} first established a minimax lower bound $\Omega(\sqrt{D O A T})$, where $D$ is the MDP diameter as defined in Section \ref{sec: setting}, and the 'state transition distributions and rewards of the MDP are assumed to be time-invariant over the time horizon $T$ under consideration ($O$ denotes the number of states and $A$ the number of actions). Based on upper confidence bounds, \cite{auer2008near} has also proposed \texttt{UCRL2}, an algorithm that achieves a regret bound of $\tilde{\mathcal{O}} (D O \sqrt{A T})$. Variants of \texttt{UCRL2} with improved regret bounds were also proposed later, but are omitted in this manuscript due to their lack of efficiency in practice, e.g., computational inefficiency as in \cite{zhang2019regret}, despite their minimax optimality. 
\subsection{Non-stationary RL, as modeled by MDPs}
In a rather naive way, \cite{UCRL2} already considers the case where a predefined \textit{known-to-agent} number of changes $\ell$ should occur in the environment. Based on this, \texttt{Restarted-UCRL2} periodically restarts \texttt{UCRL2}. More recently, a number of papers have developed \cite{SWUCRL, ortner2020variational, SWUCRL2CW} Algorithms for non-stationary RL in the tabular setting. Such algorithms assume that the MDP is constant over all episodes up to the current one, say $k$, and estimate the state transition distributions and rewards based on the data up to $k-1$. If a change occurs between episodes $k-1$ and $k$, which is generally possible, the estimator \textit{biased} and \cite{ortner2020variational} show that the algorithms suffer a linear regret that scales with the norm of the bias. As a remedy, \cite{SWUCRL, SWUCRL2CW} proposes a sliding-window approach in which estimators favor recently observed transitions and penalize older ones. As shown in \cite{SWUCRL2CW}, the \cite{SWUCRL} approach leads to sub-optimal \textit{regret} bounds.\cite{SWUCRL2CW} proposes a \textit{confidence widening} variant to \cite{SWUCRL} in which the regret bounds for \textit{smoothly-changing} MDPs are more favorable and thus only handle the case where the state transition distributions of the environment are set to change up to a \textit{variation budget} assumed at initialization. \cite{ortner2020variational} periodically restarts the algorithm and discards past data at each restart, but requires much more information about the variations in state transition distributions and rewards between restarts than \cite{SWUCRL2CW} to achieve its \textit{dynamic regret} \footnote{In contrast to ours and \cite{SWUCRL}'s, which defines regret at time $t$ as the difference between the reward achieved by the active MDPs optimal policy at time $t$ and that of the learner's policy, \cite{SWUCRL2CW,ortner2020variational} consider a different notion of regret, which they also call \textit{dynamic regret}. The latter is defined as the difference between the learner's policy and the best achievable steady-state policy. Although the term \cite{yu2009arbitrarily, even2009online, neu2010online, dick2014online} is widely used, it is generally not very useful because the best steady-state policy can still lead to undesirable rewards, especially in environments with nonsmooth change} bound. Although not directly related to the setting we consider, we would also like to highlight the contributions of \cite{yu2009arbitrarily, neu2010online, arora2012deterministic, dick2014online, jin2020learning, rivera2019large}, where state transition distributions are assumed to be fixed throughout the learning process, while rewards are allowed to change. Due to the space constraints, we choose omit other contributions that are not as directly related to our setting.
\vspace{-1em}
\subsection{Background on online Change-point detection}

In the online change-point detection literature, change-point detection algorithms are designed that allows the detection of a change in the distribution of a stochastic process from one probability distribution to another. The optimality properties in term of false alarm rate and detection delay of the algorithms are studied under several problem formulations and different model assumptions. The main theoretical foundations were set up by the work of Shiryaev \cite{shiryaev1963optimum}. Existing online change-point detection are globally categorized into two types: Bayesian approaches and non-Bayesian algorithms. The former provide uncertainty quantities of the detection while the latter mainly focuses on measuring the discrepancy of the data statistics before and after the change-point. Several Bayesian methods for online change-point detection (\cite{alami20a, bayesianUAI, knoblauch2018spatio,saatcci2010gaussian}) rely on the standard Bayesian Online Change-Point Detection (\cite{fearnhead2007line}) that recursively models the posterior probability of the elapsed time since the last change. On the other hand, non-Bayesian methods mainly rely on the likelihood ratio test \cite{severo2006change, pmlr-v98-maillard19a, page1954continuous} that also leads to false positive when the probability of the latest observations decrease given an outlier. 

\vspace{-1em}

\section{Problem Formulation}
\label{sec: setting}

An instance of an MDP can be concisely specified by the tuple $M(\sO, \sA,  P, R, T)$, where $\sO = \left\lbrace 1,..., O \right\rbrace$ represents the finite set of states ($ O= |\sO|$), $\sA = \left\lbrace 1,..., A \right\rbrace$ denotes the finite set of actions ($A = |\sA|$), $T$ is the finite time horizon and ${\{R_t\}_{t=1}^T}_{o \in \sO, a \in \sA}$ is the sequence of distribution rewards. For a given $t$ and state-action pair $(o,a)$, $r_t \sim R_t(o, a)$ is drawn i.i.d according to some unknown distribution on $[0, 1]$. Moreover, we define the sequence of state transition distributions $P=\{P_t\}_{t=1}^T$, with $P_t = \{P_t(.|o, a)\}_{o \in \sO, a \in \sA}$, where $P_t(.|o, a)$ is a \textbf{multinomial} probability distribution over $\sO$ for each state-action pair $(o,a)$ at a given time instance $t$. While constraining the state transition distributions to be generated from a multinomial distribution may seem restrictive at first glance, it enjoys a widespread interest from various research communities, from which we list a few key applications: modelling the collisions in cognitive radio, monitoring the performances of statistical models, monitoring events in probes for network supervision, the multi armed bandit problem, experiments in clinical trials and recommender systems to name a few.

To clearly define our problem of interest, we outline the main assumptions to be considered. 

\textbf{Non-Stationarity.} In the contrary to many approaches proposed by literature, we assume in all generality that the state transition distributions and rewards can change arbitrarily at unknown time steps (referred to as \textit{change-points} thereof), i.e, the changes are not constrained to accumulate to a certain predefined \textit{variation budget} for example. We also, more fundamentally, make the minimal assumption that the finite set of states and that of actions are constant throughout the learning process. Moreover, being part of the exponential family of probability distributions, it exhibits a favorable concentration behavior, allowing the smooth integration to optimistic exploration based algorithms relying on upper-confidence bounds.

\paragraph{Exogeneity.} We assume neither the change-points nor the changes in reward distributions and state transition distributions depend on the previous behavior of the algorithm or the filtration of the MDP parameter history $(o_1, a_1, r_1, ..., o_t, a_t, r_t)$. It can be assumed that the set of change-points $\{c_\ell\}_{\ell=1}^{K_T}$ is generated in advance through an exogenous process, where $K_T$ is the number of changes up to time horizon $T$.

\paragraph{Convergence \& Bounds.} For convergence, we naturally assume bounded rewards, i. e, $\abs{R_t(o, a)} \leq \max\limits_{t, o, a} R_t(o, a)  <\infty, \forall o \in \sO, \ \forall a \in \sA, \forall t \in  \left\lbrace 1,..., T \right\rbrace $. 

\paragraph{Endogeneity.} The agent starts at some arbitrary state $o_{\text{init}}$. At time $t$, they observe state $o_t \in \sO$ and take action $a_t \in \sA$ according to some policy $\pi \in \Pi$. As a result, they transition to the next state $o_{t+1} \in \sO$ according to state transition distribution $P(.|o_t, a_t)$, receiving stochastic reward $r_t \sim R_t(o_t, a_t)$ drawn according to some unknown distribution on $[0, 1]$. The endogeneity assumption here restricts the set of feasible policies $\Pi$ to be \textit{non-anticipatory}, i.e, the policy choice only depends on the current state and the set of previous observations $(o_1, a_1, r_1, ..., o_t, a_t, r_t)$. 

Now that we have introduced the set of assumptions we consider, we are now ready to formulate the definitions and that will be used throughout the paper. 

\paragraph{Switching-MDP Problem.} Following a first instance of the name for Multi-Armed Bandits \cite{garivier2011upper}, and then in \cite{SWUCRL} for MDPs, we adopt the name \textit{switching-MDP} to characterize our setting. Defining the set of change-point times as $\{c_\ell\}_{\ell=0}^{K_T}$, we consider the piecewise stationary MDP $M$ to be in configuration $M_\ell(\sO, \sA, P_\ell, R_\ell, T)$ for $t \in [c_\ell, c_{\ell+1})$. Hence the switching-MDP problem $M$ can be fully expressed by the tuple $\mathbf{M} = \{\mathbb{S}=\{M_0,..,M_{K_T-1}\}, \mathcal{C}=\{c_0,..,c_{K_T}\}\}$, where $c_0$ and $c_{K_T}$ are the respective learning start ($t=1$) and end ($t=T$) times. 



\paragraph{Diameter of a MDP.} The diameter of an MDP $M_\ell$ is  defined as follows:
\begin{center}
        $
        D\left(M_\ell\right) = \max\limits_{o_1, o_2 \in \sO, o_1 \neq o_2}\min\limits_{\pi \in \Pi} \mathbb{E}\left[\tau\left(o_1,o_2,M_\ell,\pi\right)\right] $
\end{center}

where the random variable $\tau\left(o_1,o_2,M_\ell,\pi\right)$ denotes the number of steps a policy $\pi$ from the set of feasible stationary policies $\Pi$ takes to move from $o_1$ to $o_2$ on average. In particular, we refer to an MDP with a finite  diameter as a \textit{communicating MDP}. 

As we use \textit{regret} as a performance measure throughout this article, as it is done in numerous other learning settings, we introduce the definition of the \textit{average reward} for a constituent MDP $M_\ell$ of $\mathbf{M}$, given the execution of an algorithm $\gA$ following a stationary policy, which can be written as follows: $\rho(M_\ell, \gA, o_{\text{init}}) \dfq \lim\limits_{T \to \infty} \frac{1}{T} \mathbb{E}\left[\sum\limits_{t=1}^{T}  r_t\right]$, where the sequence of rewards is assigned following the states and actions chosen by the policy generated by $\gA$ starting from $o_{\text{init}}$. 

Assuming each of the MDPs that constitute $\mathbf{M}$ to be communicating, we get by virtue of \cite{puterman2014markov} that the optimal average reward does not depend on the initial state of the MDP. This result is fundamental to the definition of the regret, as it allows to decompose the optimal average reward for $\mathbf{M}$ into the sum of that for its constituent MDPs $\{M_\ell\}_{\ell=0}^{K_T-1}$. Now, defining the optimal reward for MDP $M_\ell$ as follows: $\rho_{M_\ell}^{\star} \dfq  \max\limits_{\pi, o \in \sO} \rho(M_\ell, \gA, o)$
where, again, $\pi$ is the non-anticipatory policy generated by $\gA$.  Now we are ready to define the regret for a switching-MDP problem as follows.

\paragraph{Regret for a switching-MDP.} The regret of an algorithm $\gA$ for a switching-MDP problem $\mathbf{M} = \left(\{M_\ell\}_{\ell=0}^{K_T}, \{c_\ell\}_{\ell=1}^{K_T}\right)$ up to time horizon $T$ starting from some initial state $s$ is written as: 
\begin{center}
    $\mathfrak{R}\left(\mathbf{M}, \gA, o, T\right) = \sum\limits_{t=1}^T\left(\rho_{\mathbf{M}}^\star(t)- \esp{r_t} \right) $ \quad \text{where} \quad $\rho_{\mathbf{M}}^\star(t) \dfq \rho_{M_\ell}^\star$ if $t \in [c_\ell, c_{\ell+1})$. 
\end{center}

\section{Change-point Detection as remedy to Non-stationarity}

In this section, we start by designing the multinomial version of the Restarted Bayesian Online Change-point detector introduced in \cite{alami20a}. Then, we provide the mathematical guarantees of this algorithm in term of false alarm rate and detection delay. Finally, we design the \texttt{R-BOCPD-UCRL2} strategy, an \texttt{UCRL2} instance equipped with the \texttt{R-BOCPD} in order to handle piecewise stationaty MDPs. 

\subsection{Restarted Bayesian Online Change-point detection for multinomial distributions} 

In this section, we study the online change point detection problem, where a sequence of
independent multivariate random variables with common fluctuation upper bound are
collected, and the mean may change at one or multiple time points. Indeed, we consider an agent aiming at detecting changes \modif{in the generation} of an online stream.
At each time step $t$, the agent observes the datum $x_t \sim \multi{\mu_{1,t},..., \mu_{O,t}}$: a random variable following the multinomial distribution of parameters $\left(\mu_{1,t},..., \mu_{O,t}\right) \in \left[0,1 \right]^O$ ($x_t \in \left\lbrace 1,..., O \right\rbrace  $) and need to decide whether or not there is a change in the generation of the stream. Alternatively,  the agent may compute at each time step $t$, an estimation  $\hat{c}_t$ of the last change-point.

\paragraph{Notations.}
In the following, we denote $n_{s:t} \define t-s+1$, the number of observations from time $s$ until time $t$. Moreover,  the empirical frequency of observing $o$ in the sequence $\bfx_{s:t} = \left(x_s,...,x_t \right)$ is denoted as $\hat{\mu}_{o,s:t} \define \frac{1}{n_{s:t}} \sum_{s'=s}^t \indFct{x_{s'}= o}$.

\begin{definition}[Kullback Leibler divergence for multinomial distributions]
Let's $\btheta^{(1)}  = \left(\theta^{(1)}_1,..., \theta^{(1)}_O \right)$ and 
$\btheta^{(2)}  = \left(\theta^{(2)}_1,..., \theta^{(2)}_O \right)$
be the paramereters of two multinomial distributions, then the relative entropy from $\multi{\theta^{(2)}_1,..., \theta^{(2)}_O}$ and $\multi{\theta^{(1)}_1,..., \theta^{(1)}_O}$ is defined as follows: $\infdiv*{\btheta^{(2)}}{\btheta^{(1)}} = \sum_{o=1}^{O} \theta^{(2)}_o  \log \frac{\theta^{(2)}_o}{\theta^{(1)}_o}$.
\end{definition}

\begin{definition}[Piecewise stationary multinomial process]
	Let $T$ denote the stream length and $K_T$ the overall number of change-points observed until time $T$.
	We assume that the observations $x_t \sim \multi{\mu_{1,t},..., \mu_{O,t}}$ are generated by a piecewise multinomial process such that there exists a non-decreasing change-points sequence $\left( c_\ell  \right)_{\ell \in \seg{1,K_T} } \in \mathbb{N}^{K_T}$ verifying:
	\vspace{-1mm}\noindent
	\begin{align}
	\forall  \ell \in  \left\lbrace 0,...,K_T \right\rbrace, \  \forall  t \in \left[ c_\ell, c_{\ell+1} \right), \forall o \in \sO \quad  \mu_{o,t} = \theta_{o,\ell} ,
	\quad \text{where:} \
	c_0 = 1 < c_2 <...<c_{K_T} = T. &  
	\label{PwSBEnv}
	\end{align}
\end{definition}

\begin{theorem}[Lower Bound for the detection delay]
Let: $\left( x_r,...,x_{c_\ell-1} \right) \sim \multi{\btheta^{(l-1)} = \left(\theta^{(\ell-1)}_1,..., \theta^{(\ell-1)}_O \right)}$ and $\left(x_{c_\ell},...,x_{t}\right) \sim \multi{\btheta^{(\ell)} = \left(\theta^{(\ell)}_1,..., \theta^{(\ell)}_O\right) }$, $\pA$ an online change-point detection strategy, $c_\ell$ the change-point to detect and $r$ the restarting time.
Assuming that the false alarm rate is controlled such that: $\Pr{\exists s \in \segOR{r,\tau_c}: \pA \OpSeg{\bfx_{r:s}}=1}{\btheta^{(\ell-1)}} \leq \delta$, then as the quantity $\frac{n_{r:c_\ell}}{\abs{\log \delta}} \underset{\delta \rightarrow 0}{\rightarrow} \infty$, 
the expected detection delay $\Esp{\StoppingTime{\pA}{\bfx_{r:t}} - c_\ell}{\btheta^{(\ell-1)},\btheta^{(\ell)}}$ is lower bounded as follows: 
$
\Esp{\StoppingTime{\pA}{\bfx_{r:t}} - c_\ell}{\btheta^{(\ell-1)},\btheta^{(\ell)}} \geq  \OpSeg{\frac{\Pr{\StoppingTime{\pA}{\bfx_{r:t}} > c_\ell}{\btheta^{(\ell-1)}}}{\infdiv*{\btheta^{(\ell)}}{\btheta^{(\ell-1)}}}} \log \frac{1}{\delta}$.
\label{Theorem::LowerBoundDetectionDelay}
\end{theorem}

\paragraph{Extension of the Restarted Bayesian Online Change-point detector (R-BOCPD) for multinomial distributions.}

\cite{alami20a} introduced Restarted Bayesian Online Change Point Detection (\texttt{R-BOCPD}),
which is a pruned version of Bayesian Online Change-point Detector applicable for univariate Bernoulli-distributed samples
with changes in the mean of the distribution.

In this section, we propose to extend the \texttt{R-BOCPD} algorithm introduced in \cite{alami20a} in order to deal with multinomial distributions. First, we start by extending the Laplace predictor for multinomial distributions.

\begin{definition}[Extending the Laplace predictor]
    \label{LaplacePred}
	The predictor $\predictor{x_{t+1}}{\bfx_{s:t}}$ takes as input a sequence $\bfx_{s:t} \in \event{1,..., O}^{n_{s:t}}$  and predicts the value of the next observation $x_{t+1} \in \event{1,..., O}$ as follows
	\begin{align}
	\predictor{x_{t+1}}{\bfx_{s:t}}  \define 
	\frac{\sum_{i=s}^{t} \indFct{x_i = x_{t+1}}   + 1 }{n_{s:t} + O } 
	\label{Eq::Forecaster}
  \end{align}
		\label{Def_LapPredic}
where $\forall  x \in \event{1,..., O} \ \predictor{x}{\emptyset}  = \frac{1}{O}$ corresponds to the uniform prior given to the process generating  $\theta_c$.
\end{definition}

Recall that in the work of \cite{alami20a}, instead of dealing with a run-length, they deal with the notion of forecaster that is a product of successive Laplace predictors. Thus, as in \cite{alami20a}, we introduce the loss of a forecaster. 

\begin{definition}[Forecaster loss]
Using the predictor, the instantaneous loss of the forecaster $s$ at time $t$ is given by:
\begin{align*}
 l_{s:t} &\define 
 -\log \predictor{x_{t}}{\bfx_{s:t-1}} = - \sum_{o=1}^O \indFct{x_{t} = o}  \log \predictor{o}{\bfx_{s:t-1}}.
\end{align*}
Then, let $\hat{L}_{s:t}\define \sum_{s'=s}^t l_{s':t}$ denotes the cumulative loss incurred by the forecaster $s$ from time $s$ until time $t$ which takes the following crude expression:
\begin{align}
    \hat{L}_{s:t}\define \sum_{s'=s}^t -\log \predictor{x_{t}}{\bfx_{s':t-1}}
    \label{Eq::cumulLoss}
\end{align}
    
\end{definition}

Thus, the forecaster weights update will remain the same (for some temporal function $\eta_{r,s,t}\in \OpSeg{0,1}$).
\begin{align}
    \omega_{r,s:t} & = \begin{cases}
\frac{\blue{\eta_{r,s,t}}}{\blue{\eta_{r,s,t-1}}}	 \exp \left(-l_{s,t} \right) \omega_{r,s:t-1} & \forall s<t,\\
    \eta_{r,t,t} \times \pW_{r:t-1} & \text{}s=t\,.
    \label{eq::weight}
\end{cases} \text{with} \quad \pW_{r:s-1} \define  \exp \left(-\hat{L}_{r:s-1} \right) \ \text{for some starting time $r$.}
\end{align}

Finally, we keep the same restart procedure as in \cite{alami20a}:

\begin{align}
\restart{x_r,...,x_t} =  \indFct{ \exists s \in \segOL{r,t}:\omega_{r,s:t} >  \omega_{r,r:t}}
\label{eq::Restart}
\end{align}

We describe the \texttt{R-BOCPD} algorithm for multinomial distributions in Algorithm \ref{alg:R-BOCPD}.

\begin{algorithm}[H]
	\caption{\texttt{R-BOCPD} for multinomial distributions} \label{alg:R-BOCPD}
	\begin{algorithmic}[1]
		\REQUIRE $\eta_{r,s,t}\in \OpSeg{0,1}$
     \STATE $r \leftarrow 1$, $\omega_{r,1:1} \leftarrow 1$, $\eta_{r,1,1} \leftarrow 1$.
		\FOR{$t=1,\dots$}
		\STATE Observe $x_t \sim \multi{\mu_{1,t},..., \mu_{O,t}}$
		\STATE Define for each forecaster $s$ from time $r$ to time $t$:
		$
		\omega_{r,s:t}  \leftarrow \begin{cases}
		\frac{\eta_{r,s,t}}{\eta_{r,s,t-1}}	 \exp \left(-l_{s:t} \right) \omega_{r,s:t-1} & \forall s<t,\\
			\eta_{r,t,t} \times \pW_{r:t-1} & s=t\,.
		\end{cases}
        $
        \STATE \text{\bf if} {$\restart{x_r,...,x_t} = 1$} \text{\bf then} $r \leftarrow t+1$, $\omega_{r,r:r} \leftarrow 1$, $\eta_{r,r,r} \leftarrow 1$. 
		\STATE Estimate the last change-point: $\hat{c}_t \leftarrow r$.
		\ENDFOR
	\end{algorithmic}
\end{algorithm}

\subsection{Performance Guarantees for the \texttt{R-BOCPD} in the multinomial case}
In this section, we provide sufficient conditions on the parameter $\eta_{r,s,t}$ that guarantee the false alarm rate control (Theorem \ref{Theorem:FalseAlarmRate}) and the finite detection delay (Theorem \ref{Theorem:DetectionDelay}) for the \texttt{R-BOCPD} algorithm in the multinomial case.

\subsubsection{Controlled False-alarm Rate}
\begin{theorem}[False-alarm rate]
Let: $\boldsymbol{\theta}  = \left(\theta_1,..., \theta_O \right)$ denotes the vector of the parameters for a Multinomial distribution $\multi{\theta_1,..., \theta_O}$. For $r < t$, assume that $\left(x_r, ..., x_t\right) \sim \multi{\theta_1,..., \theta_O}^{\otimes n_{r:t}}$. Let $\alpha > 1$, if $\eta_{r,s,t}$  is small enough such that:

\begin{align}
\eta_{r,s,t} &< \left( \prod_{i=1}^{O-1} \frac{\left( n_{r:s-1}+i  \right)\left( n_{s:t}+i  \right)}{n_{r:t}+i} \right) \times \frac{\exp{(2b_1)}}{\left( n_{r:s-1} n_{s:t} \right)^{\frac{O-1}{2}} \times (O-1)! } \times \left(\frac{\log(4\alpha+2)\delta^2}{4n_{r:t}\log(\left( \alpha+3 \right) n_{r:t})} \right)^\alpha \label{Eq::UBEta} \\
&\text{where: } b_1 = -\frac{O}{12}- \frac{O-1}{2} \log \left(2\pi \right) + \frac{O}{2} \log O. \nn
\end{align}

	then, with a probability higher than $1-\delta$, no false alarm occurs on the interval $\left[r, c_\ell \right) $:
	\begin{align*}
	\Pr{\exists\  t \in \segOR{r, c_\ell}: \restart{x_r,...,x_t} = 1 }{\btheta} \leq  \delta.
	\end{align*}
	\label{Theorem:FalseAlarmRate}
\end{theorem}

\begin{definition}[Relative gap $\Delta_{o,r,s,t}$]
Let $\Delta_o \in \seg{0,1}$. The relative gap $\Delta_{o,r,s,t}$ for the forecaster $s$ at time $t$ takes the following form (depending on the position of $s$):
    \begin{align}
	\Delta_{o,r,s,t} = \left( \frac{ n_{r:c_\ell-1}}{ n_{r:s-1}} \indFct{c_\ell\leq s \leq t} + \frac{n_{c_\ell:t}}{ n_{s:t}} \indFct{s < c_\ell} \right)\Delta_o. \nn
	\end{align}
\end{definition}

\vspace{-1em}

\subsubsection{Optimal Detection Delay}

\begin{theorem}[Finite detection delay]
Let $\left( x_r,...,x_{c_\ell-1} \right) \sim \multi{\theta^{(\ell-1)}_1,..., \theta^{(\ell-1)}_O}^{\otimes n_{r:c_\ell-1}}$, $\left(x_{c_\ell},...,x_{t}\right) \sim \multi{\theta^{(\ell)}_1,..., \theta^{(\ell)}_O}^{\otimes n_{c_\ell:t}}$, $\Delta_o = \abs{\theta^{(\ell-1)}_o - \theta^{(\ell)}_o}$: the change-point gap related to observation $o 
\in \left\lbrace 1,..., O \right\rbrace$ and $\bDelta = \left(\Delta_1,...,\Delta_O \right)$ the vector of change-point gap. Then, let: $f_{r,s,t} = \sum_{i=1}^{O-1} \log \left( n_{r:s-1}+i  \right) + \sum_{i=1}^{O-1} \log \left( \frac{n_{s:t}+i}{n_{r:t}+i}  \right) - \frac{O-1}{2} \log \left( \frac{n_{s:t}}{n_{r:t}} \right) - \log(O-1)! $. If $\eta_{r,s,t}$ is large enough such that:
\begin{align}
\eta_{r,s,t} >  
\exp\big(-2n_{r,s-1}\sum_{o=1}^O\left(\Delta_{o,r,s,t} - \pC_{r,s,t,\delta}  \right)^2 + f_{r,s,t} \big), 
\label{Eq::LWEta}
\end{align}
then, the change-point $c_\ell$ is detected (with probability at least $1-\delta$) with delay not exceeding $	\kD_{\bDelta,r,c_\ell}$, such that:
\begin{align}
&\kD_{\bDelta,r,c_\ell} =  \min \Eventt{d \in \mathbb{N}^\star: d > \frac{2}{\sum_{o=1}^{O}\left(\Delta_o - \pC_{r,c_\ell, d+ c_\ell-1 ,\delta}  \right)^2} \times \frac{-\log \eta_{r,c_\ell,d+c_\ell-1} + f_{r,c_\ell,d+c_\ell-1}}{1+\frac{2\left(\log \eta_{r,c_\ell,d+c_\ell-1} - f_{r,c_\ell,d+c_\ell-1}\right)}{n_{r,c_\ell-1}\times   \textcolor{black}{\sum_{o=1}^{O}\left(\Delta_o - \pC_{r,c_\ell, d+ c_\ell-1 ,\delta}  \right)^2}}}} \\
&\text{where: } \pC_{r,s,t,\delta} = \frac{\sqrt{2}}{2} \Bigg( \sqrt{\frac{1+\frac{1}{n_{r:s-1}}}{n_{r:s-1}} \log \left(\frac{2\sqrt{n_{r:s}}}{\delta}\right)} +\sqrt{\frac{1+\frac{1}{n_{s:t}}}{n_{s:t}} \log \left(\frac{2n_{r:t} \sqrt{n_{s:t}+1} \log^{2}\left(n_{r:t}\right)}{\log(2) \delta}\right)}\Bigg). \label{Eq::C}
\end{align}
\label{Theorem:DetectionDelay}
\end{theorem}

\begin{discussion}({Asymptotic behavior of the detection delay)}
The asymptotic regime corresponds to the case where the elapsed time between the last restart $r$ and the new change point $c_\ell$ tends to infinity, while the probability of false alarm $\delta$ tends to zero. Thus, we get:

\begin{align}
\kD_{\bDelta,r,c_\ell}  \underset{n_{r,c_\ell-1} \rightarrow \infty}{\rightarrow}
\  \frac{2 \left( -\log \eta_{r,\tau_c,d+ \tau_c-1}+ \smallO{\log \tfrac{1}{\delta}}\right)}{\sum_{o=1}^{O}\Delta_o^2} \overset{(a)}{=}  \order{\frac{\log \frac{1}{\delta}}{\infdiv*{\btheta^{(\ell)}}{\btheta^{(\ell-1)}}} }
\label{eq::DetectionDelay}
\end{align}    
\end{discussion}

where (a) originates from the Pinsker inequality that relates the Kullback-Leibler divergence to the total variation. Thus, following the statement of Theorem \ref{Theorem::LowerBoundDetectionDelay}, the detection delay $\kD_{\bDelta,r,c_\ell}$ of the \texttt{R-BOCPD} is asymptotically order optimal.

\begin{discussion}{(Choice of the hyperparameter $\eta_{r,s,t}$)}
    In order to guarantee the false alarm rate and detection delay, we need to choose an appropriate form of the parameter $\eta_{r,s,t}$ that satisfies both conditions in Equation (\ref{Eq::LWEta}) and Equation (\ref{Eq::UBEta}). Choosing $\eta_{r,s,t} \approx \frac{1}{n_{r:t}}$ is satisfying both conditions.
\end{discussion}

\subsection{The \texttt{R-BOCPD-UCRL2} Strategy}

Now, we propose to equip \texttt{UCRL2} in the switching-MDP setting $\mathbf{M}$ with \texttt{R-BOCPD}. This originates from the ability  
to decompose $\mathbf{M} = \{\mathbb{S}=\{M_0,..,M_{K_T-1}\}, \mathcal{C}=\{c_0,..,c_{K_T}\}\}$ according to independently generated stationary periods $(c_\ell, c_{\ell+1})$, along which \texttt{R-BOCPD} first detects the switch from MDP $M_\ell$ to $M_{\ell+1}$ at change instance $c_\ell$ and restarts $\texttt{UCRL2}$ accordingly with minimal delay with high probability as quantified by the provided upper confidence bound. We also ensure that no other restarts occur during a stationary period with high probability in a similar way. The exact approach is explained in more detail in Algorithm \ref{alg:R-UCRL}.

\begin{definition}{(\texttt{UCRL2} Framework).}
We adopt the same notation and learning framework as in the original \texttt{UCRL2} \cite{auer2008near}, which we omit here due to space constraints. We list that in more depth in Appendix \ref{sec::framework}. 
\end{definition}

\begin{theorem}[Finite-time Optimal Regret Upper Bound] With probability at least $1-\delta$, it holds for a switching-MDP problem $\mathbf{M} = \{\mathbb{S}=\{M_0,..,M_{K_T-1}\}, \mathcal{C}=\{c_0,..,c_{K_T}\}\}$ (starting at some initial state $o_{c_0}$) with (piecewise) stationary periods of length at least 1 that the \texttt{R-BOCPD-UCRL2} regret as defined in Section \ref{sec: setting} is bounded as follows:
\begin{align*}
    \mathfrak{R} \left(\mathbf{M}, \texttt{R-BOCPD-UCRL2}, o_{c_0}, T\right) \leq 34 D O \sqrt{A T K_T \log\left (\frac{T}{\delta} \right)} + \sum\limits_{\ell=0}^{K_T-1} \kD_{\bDelta_{\ell+1},c_{\ell}+d_{\ell},c_{\ell+1}} 
\end{align*}
where $\kD_{\bDelta_{\ell+1},c_{\ell}+d_{\ell},c_{\ell+1}}$ is \texttt{R-BOCPD}'s detection delay on input stream $(o_{c_{\ell}+d_{\ell}}, ..., o_{c_{\ell+1}})$ for the gap $\bDelta_{\ell+1} =  \left(\Delta_{1,\ell+1},...,\Delta_{O,\ell+1} \right) \quad \text{where} \quad \Delta_{o,\ell+1} =  \abs{\theta^{(\ell)}_o - \theta^{(\ell+1)}_o}$, with $\btheta^{(\ell)} = \left(\theta^{(\ell)}_1,...,\theta^{(\ell)}_O \right)$ and $\btheta^{(\ell+1)} = \left(\theta^{(\ell+1)}_1,...,\theta^{(\ell+1)}_O \right)$ being the pre and post state-transition kernels over the set of states $\sO$ for change-point $c_{\ell+1}$.
\label{main_theorem}
\end{theorem}

Now, we introduce a corollary to characterize the asymptotic behavior of \texttt{R-BOCPD-UCRL2}'s regret.

\begin{corollary}[Asymptotic Regret Upper Bound]
With probability at least $1- \delta$, assuming $c_{\ell+1}-c_\ell-d_\ell \gg 1$ with false-alarm probability $\delta_{\text{False-Alarm}} \to 0$ (as in Equation (\ref{eq::DetectionDelay})), we can write the asymptotic upper bound for the regret of \texttt{R-BOCPD-UCRL2} on $\mathbf{M}$ starting at some state $o_{c_0}$ as follows:
\begin{align*}
    \mathfrak{R} \left(\mathbf{M}, \texttt{R-BOCPD-UCRL2}, o_{c_0}, T\right) = \order{D O \sqrt{A T K_T \log\left (\frac{T}{\delta} \right)} + \frac{K_T \log \frac{K_T}{\delta}}{\min\limits_\ell \: \infdiv*{\btheta^{(\ell+1)}}{\btheta^{(\ell)}}}}
\end{align*}
\label{main_cor} 
\end{corollary}
\begin{discussion}{(Optimality of the Upper Bound)}
We derived both a finite-time and an asymptotic variant of \texttt{R-BOCPD-UCRL2}'s regret upper bound, both comparing favorably to state-of-the-art. Given the purpose of design of \texttt{R-BOCPD-UCRL2}, which is to allow it to adapt to rapidly and abruptly changing non-stationary RL environments, the regret bound correlates optimally with the distance between the distributions before and after the change-point. We also highlight that our proposed approach is the first one to obtain a regret of $\tilde{\mathcal{O}}(T^{\frac{1}{2}})$ up to our knowledge. Previously proposed sliding-window approaches \cite{SWUCRL, SWUCRL2CW}  obtain a regret bound of $\tilde{\mathcal{O}}(T^{\frac{2}{3}})$ and $\tilde{\mathcal{O}}(T^{\frac{3}{4}})$ respectively. \texttt{UCRL2} with restarts \cite{UCRL2}, even while restarting $T^{\frac{1}{3}}K_T^{\frac{-1}{3}}$ more times than \texttt{R-BOCPD-UCRL2}, still only obtains a regret of $\tilde{\mathcal{O}}(T^{\frac{2}{3}})$.
\end{discussion}

\vspace{-1.4em}

\section{Experiments} \label{xps:main}
\vspace{-1em}
We benchmark \texttt{R-BOCPD-UCRL2} against $4$ algorithms that perform the best within our setting, to the best of our knowledge. We list them as follows:
\begin{itemize}
    \item \textbf{Sliding-Window UCRL2} (\texttt{SWUCRL2},  \cite{SWUCRL}): uses a sliding-window approach to only maintain the last $W$ time steps of the filtration history, where $W$ is referred to as the \textit{window-size}. 
    \item \textbf{Sliding-Window UCRL2 with Confidence Widening} (\texttt{SWUCRL2-CW}, \cite{SWUCRL2CW}) rely on even more \textit{optimism} than \texttt{SWUCRL2}, where in addition to a window of size $W$, defines a \textit{confidence widening} parameter $\eta$ that quantifies the amount of additional optimistic exploration to be done on top of the conventional optimistic exploration realized via upper confidence bounds. 
    \item \textbf{Restarted UCRL2} (\texttt{Restarted-UCRL2}, \cite{UCRL2}), define a variant of vanilla \texttt{UCRL2} where  the latter is restarted at steps $\tau_i = \ceil{\frac{i^3}{{K_T}^2}}$ and where the number of changes $K_T$ is assumed to be known at initialization. 
    \item \textbf{Vanilla UCRL2} (\texttt{UCRL2}, \cite{UCRL2}).
\end{itemize}
 
 Moreover, we consider a variant of \texttt{UCRL2}, which is regret-optimal as per our regret definition in Section \ref{sec: setting} (as \texttt{UCRL2} is near-optimal at each stationary period $[c_\ell, c_{\ell+1})$). It is defined as follows:
 \begin{itemize}
     \item \textbf{Oracle-Equipped UCRL2} (\texttt{UCRL2 Oracle}) is \textit{aware} of all the changes $\left\{c_\ell\right\}_{\ell=1}^{K_T-1}$ already at initialization and hence restarts \texttt{UCRL2} exactly at each $c_\ell$ for $\ell \in \left\lbrace 0,..., K_T-1 \right\rbrace$. 
 \end{itemize}
 
 We list the exact experimental setup along with the hyperparameters of choice of each algorithm in more detail in Appendix \ref{sec:xps}.  


\subsection{Experimental Results}
We evaluate the performance of the aforementioned algorithms on a variety of synthetically generated MDPs, with state-action sets of different cardinalities. The change-points are (randomly) generated up to time horizon $T$, allowing to examine the effect of changing the duration in-between change-points on learning. We plot the cumulative rewards of each approach as a function of time as follows

\begin{figure}[H]
    \vspace{-1.2cm}
    \hspace*{-2.25cm}   
    \includegraphics[scale=.75]{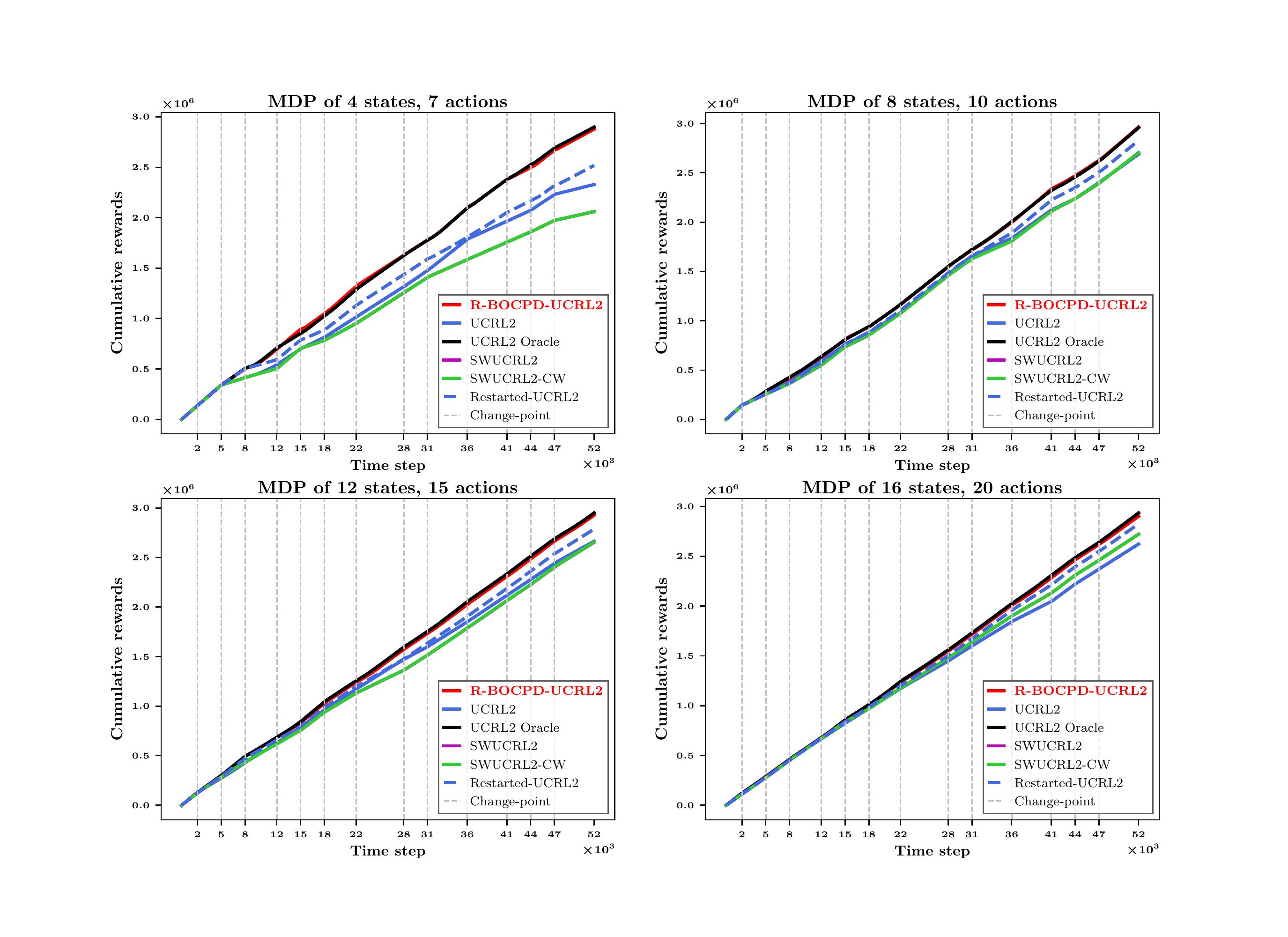}
    \vspace{-1.7cm}
    \caption{Benchmark of \texttt{R-BOCPD-UCRL2} against state-of-the-art for various state-action pairs for a sequence of random change-points. The level of abruptness, i.e the variation of the state-transition distributions and rewards also varies among change-points, allowing to model both globally and locally induced changes to the MDPs.}
    \label{fig:benchmark}
\end{figure}
\vspace{-.5cm}
\subsection{Performance Evaluation}
Figure \ref{fig:benchmark} shows that \texttt{R-BOCPD-UCRL2} is (nearly) regret-optimal in practice, as its performance in the defined general setting is very close to that of the change-point aware (optimal) \texttt{UCRL2 Oracle}. We also highlight that the observed performance generalizes well beyond switching-MDP problems of various state and action space sizes and different total \textit{variation budgets} for both the state-transition distributions and rewards. Given the space constraints, a more in-depth discussion of the performance of each algorithm along with its key assumptions is provided in Appendix \ref{sec:xps_discussion}. 
\section{Discussion \& Overall Remarks}
In this work, we proposed the Restarted Bayesian Online Change-Point Detection algorithm (\texttt{R-BOCPD-UCRL2}), which is a change-point detector equipped model-based RL algorithm operating on non-stationary environments that can be fully characterized via a discrete-time piecewise-stationary MDP. We extended the theoretical guarantees of the Bayesian Online Change-Point Detection algorithm (\texttt{BOCPD}) to the more general multinomial distribution and proved that \texttt{R-BOCPD-UCRL2} is regret-optimal with an asymptotic regret bound of $\order{D O \sqrt{A T K_T \log\left (\frac{T}{\delta} \right)} + \frac{K_T \log \frac{K_T}{\delta}}{\min\limits_\ell \: \infdiv*{\btheta^{(\ell+1)}}{\btheta^{(\ell)}}}}$, which is the first to achieve a bound of $\tilde{O}(T^{\frac{1}{2}})$ in the time horizon $T$ up to our knowledge. We further proved the optimality of \texttt{R-BOCPD-UCRL2} in practice, as it compares to state-of-the-art, with much fewer restarts and no implicitly defined input parameters (MDP diameter, variation budget etc).

\textbf{Limitations and Future Work:} Here, we highlight a few ideas that were not considered within the scope of this manuscript, but still would be very promising directions in the authors' opinion. First, we address the assumption that the state-transition distributions originate from a multinomial distribution. We note that the Bayesian Online Change-point Detector does not necessarily require an input stream stemming from a multinomial distributions, but can be extended to arbitrary distributions from the exponential family of probability distributions. Given the latter, the theoretical guarantees in terms of minimal detection delay and false-alarm rate do extend as well. Next, we highlight that our algorithm is biased towards detecting change-points around which distributions change in a rather radical way, i.e with large enough variation in the sense of a total variation norm for instance. Here, a good direction would be to define a threshold value to allow the change-point detector to decide when to restart the stationary RL algorithm (here \texttt{UCRL2}) given the global/local nature of the changes to the MDP parameters. Finally, a natural extension would also be to propose new change-point detector equipped model-free non-stationary RL algorithms.    

\begin{algorithm}[H]
	\caption{\texttt{R-BOCPD-UCRL2}} \label{alg:R-UCRL}
	\begin{algorithmic}[1]
		\REQUIRE A confidence parameter $\delta \in (0,1)$,  $\eta_{r,s,t}\in \OpSeg{0,1}$
    \STATE Set $\forall (o,a)$ $N_{0}(o, a) \leftarrow 0, V_{0}(o, a) \leftarrow 0, t \leftarrow 1, k \leftarrow 1$ and observe initial state $s_{1}$.
    \STATE \textbf{Initialize restart time} $r \leftarrow 1 $
    \FOR{each $(o,a)$ pair}
		\STATE Initialize a $\textbf{R-BOCPD}_{o,a}$ procedure
	\ENDFOR

     \FOR{episodes $k \geq 1$}
     \STATE \textbf{Initialize episode $k$:}
		\STATE \Space Set the start time of episode $k$, $t_{k}=t$
        \STATE \Space For all $(o, a) \in \sO \times \sA$ initialize the state-action counts for episode $k, V_k(o, a):=0$  . Further, set the number of times any action action $a$ was executed in state $o$ prior to episode $k$ for all the states $o \in \mathcal{O}$ and actions $a \in \mathcal{A}$,
$$
N_k(o, a):=\#\left\{r \leq t < t_k: o_t=o, a_t=a\right\} .
$$

        \STATE \Space For all $o, o^{\prime} \in \sO$ and $a \in \sA$, set the observed cumulative rewards when action $a$ was executed in state $o$ and the number of times that resulted into the next state being $o^{\prime}$ prior to episode $k$,
$$
\begin{gathered}
R_k(o, a):=\sum_{t=r}^{t_k-1} r_t \indFct{o_t=o, a_t=a }, \text{and} \ P_k\left(o, a, o^{\prime}\right):=\#\left\{r \leq t <t_k: o_t = o, a_t = a, o_{t+1}=o^{\prime}\right\} .
\end{gathered}
$$
\STATE \Space \text { Compute estimates } $\hat{R}_k(o, a):=\frac{R_k(o, a)}{\max \left\{1, N_k(o, a)\right\}}, \hat{P}_k\left(o^{\prime} \mid o, a\right):=\frac{P_k\left(o, a, o^{\prime}\right)}{\max \left\{1, N_k(o, a)\right\}}$
\STATE \textbf{Compute policy $\tilde{\pi}_k$:}
\STATE \Space Let $\mathcal{M}_k$ be the set of all MDPs with state space $\sO$ and action space $\sA$, and with transition probabilities $\tilde{P}(\cdot \mid o, a)$ close to $\hat{P}_k(\cdot \mid o, a)$, and rewards $\tilde{R}(o, a) \in[0,1]$ close to $\hat{R}_k(o, a)$, such that:
$$
\begin{aligned}
\left|\tilde{R}(o, a)-\hat{R}_k(o, a)\right| & \leq \sqrt{\frac{7 \log \left(2 O A t_k / \delta\right)}{2 \max \left\{1, N_k(o, a)\right\}}} \text { and } \left\|\tilde{P}(\cdot \mid o, a)-\hat{P}_k(\cdot \mid o, a)\right\|_1 & \leq \sqrt{\frac{14 O \log \left(2 A t_k / \delta\right)}{\max \left\{1, N_k(o, a)\right\}}} .
\end{aligned}
$$
\STATE \Space Use extended value iteration to find a policy $\tilde{\pi}_k$ and an optimistic MDP $\bar{M}_k \in \mathcal{M}_k$ such that
$$
\tilde{\rho}_k:=\min_{o \in \mathcal{O}} \rho\left(\bar{M}_k, \tilde{\pi}_k, o\right) \geq \max _{M^{\prime} \in \mathcal{M}_k, \pi, o^{\prime}} \rho\left(M^{\prime}, \pi, o^{\prime}\right)-\frac{1}{\sqrt{t_k}}
$$
\STATE \textbf{Execute policy $\tilde{\pi}_k$:}
\Space \WHILE{$V_k\left(o_t, \tilde{\pi}_k\left(o_t\right)\right)<\max \left\{1, N_k\left(o_t, \tilde{\pi}_k\left(o_t\right)\right)\right\}$}
\STATE Choose action $a_t=\tilde{\pi}_k\left(o_t\right)$, obtain reward $r_t$ and observe next state $o_{t+1}$. 
\STATE Update $V_k\left(o_t, a_t\right):=V_k\left(o_t, a_t\right)+1$
\STATE Set $t:=t+1$
\STATE Perform a change-point detection test over the sequence $\left( o_r,...,o_t \right)$. \IF{$\textbf{R-BOCPD}_{o_t,a_t}.\restart{o_r,...,o_t} = 1$}  
        \STATE $\forall (o,a)$ $N_{k}(o, a) \leftarrow 0, V_{k}(o, a) \leftarrow 0, r \leftarrow t+1$.
        \ENDIF
\ENDWHILE
	\ENDFOR
	\end{algorithmic}
\end{algorithm}

\newpage

\bibliography{collas2023_conference}
\bibliographystyle{collas2023_conference}

\newpage
\appendix
\onecolumn

\appendix

\section{\texttt{UCRL2} Framework}\label{sec::framework}
We introduce our learning framework and relevant notation as follows:
\begin{itemize}
    \item $N_{k}(o, a)$ is the number of times any action action $a$ was executed in state $o$ up to episode $k$ for all the states $o \in \sO$ and actions $a \in \sA$. 
    \item  $V_{k}(o, a)$ is the number of visits to state-action pair $(o, a)$ up to episode $k$. 
    \item $P_k\left(o, a, o^{\prime}\right)$ is the state-transition kernel where action $a \in \sA$ taken at state $o \in \sO$ takes the agent to state $o^{\prime} \in \sO$, where $P_k$ is defined up to time $t_k$ starting from last \texttt{UCLR2} restart time $r$. 
    \item $\hat{P}_k\left(o^{\prime} \mid o, a\right)$ is the estimated state-transition kernel for triple $(o, a, o^{\prime})$ from the last $t_k-r+1$ observations. 
    \item $R_k(o, a)$ are the mean rewards for state-action pair $(o, a)$ up to time $t_k$ starting from last \texttt{UCLR2} restart time $r$. 
    \item $\hat{R}_k(o, a)$ are the estimated mean rewards for state-action pair $(o, a)$ given the last $t_k-r+1$ observations.
    \item $\mathcal{M}_k$ is defined as the set of statistically plausible MDPs given $\hat{P}_k\left(o^{\prime} \mid o, a\right)$ and $\hat{R}_k(o, a)$, with state space $\sO$ and action space $\sA$. 
    \item $\bar{M}_k$ is an optimistic MDP chosen from $\mathcal{M}_k$.
    \item $\tilde{P}(\cdot \mid o, a)$ is the transition kernel of $\bar{M}_k$ that is close to $\hat{P}_k(\cdot \mid o, a)$.
    \item $\tilde{R}(o, a)$ are the mean rewards of $\bar{M}_k$ that are close to $\hat{R}_k(o, a)$.
    \item $\tilde{\pi}_k$ is a near optimal policy for $\bar{M}_k$ chosen via extended value iteration, as defined in \cite{auer2008near}.
\end{itemize}    

\section{Control of the cumulative loss in the multinomial case}

\subsection{Notation and useful definitions}

	In the following, we denote by $\Sigma_{o,s:t}$ the number of times the realization $o \in \left\lbrace 1,...O \right\rbrace$ has been observed in the sequence $\bfx_{s:t}$ such that:  
	\begin{align*}
	\Sigma_{o,s:t} = \sum_{s'=s}^t \indFct{x_{s'}= o} 
	\end{align*}

\subsection{Cumulative loss close form}

Notice that: 

\begin{align*}
\forall  \bfx_{s:t} \in \left\lbrace 1,..., O \right\rbrace^{n_{s:t}} \quad \hat{L}_{s:t} \define \sum_{s'=s}^t -\log \predictor{x_{t}}{\bfx_{s':t-1}} = -\log \prod_{s'=s}^t \predictor{x_{t}}{\bfx_{s':t-1}}    
\end{align*}

Let's show by induction on $n_{s:t} \in \mathbb{N}^\star$ that: 

\begin{align}
\forall  \bfx_{s:t} \in \left\lbrace 1,..., O \right\rbrace^{n_{s:t}} \quad  \prod_{s'=s}^t  \predictor{x_{t}}{\bfx_{s':t-1}} & = \textcolor{black}{\frac{ (O-1)! }{ \left(\prod\limits_{i=1}^{O-1} \left( n_{s:t}+i \right) \right)}} \times \frac{\prod\limits_{o=1}^{O}  \Sigma_{o,s:t}! }{   n_{s:t}!}
\label{eq::rec}
\end{align}
\newpage
\begin{myproof}{of Equation (\ref{eq::rec})}
\paragraph{Step 1:} For $n_{s:t} = 1$. It means that $t=s$, $x_t \in \left\lbrace 1,..., O \right\rbrace$ and $\bfx_{s':t-1} = \emptyset$. Using the definition of the predictor $\predictor{}{}$, we obtain

\begin{align*}
    	\predictor{x_t}{\emptyset}  \define 
	\frac{ 1 }{ O}  = \textcolor{black}{\frac{ (O-1)! }{ \left(\prod\limits_{i=1}^{O-1} \left( 1+i \right) \right)}} \times \frac{\prod\limits_{o=1}^{O}  1 }{   1!}   
\end{align*}

\paragraph{Step 2:}
Assume that for some $n_{s:t} \in \mathbb{N}^\star$ that corresponds to the sequence $\bfx_{s:t} \in \left\lbrace 1,..., O \right\rbrace$, we have

\begin{align}
    \prod_{s'=s}^t  \predictor{x_{t}}{\bfx_{s':t-1}} &=  \frac{ (O-1)!
 \prod\limits_{o=1}^{O}  \Sigma_{o,s:t}! }{\left(n_{s:t}+O-1 \right)!} =  \frac{ (O-1)!
  \prod\limits_{o=1}^{O}  \Sigma_{o,s:t}! }{ \left(\prod\limits_{i=1}^{O-1} \left( n_{s:t}+i \right) \right)  n_{s:t}!}
  \label{Eq::Rec}
\end{align}

Then, observe that:
\begin{align*}
    \prod_{s'=s}^{t+1}  \predictor{x_{t+1}}{\bfx_{s':t}} = \prod_{s'=s+1}^{t+1}  \predictor{x_{t+1}}{\bfx_{s':t}} \times \predictor{x_{t+1}}{\bfx_{s:t}}
\end{align*}

Then, using the definition of the forecaster in Equation (\ref{Eq::Forecaster}) and the statement of Equation (\ref{Eq::Rec}), we obtain (for $x_{t+1} = a \in \left\lbrace 1,..., O \right\rbrace$)

\begin{align*}
    \prod_{s'=s}^{t+1}  \predictor{x_{t+1}}{\bfx_{s':t}} &= \frac{ (O-1)!
  \prod\limits_{o=1}^{O}  \Sigma_{o,s+1:t+1}! }{ \left(\prod\limits_{i=1}^{O-1} \left( n_{s+1:t+1}+i \right) \right)  n_{s+1:t+1}!} \times \frac{\sum_{i=s}^{t} \indFct{x_i = x_{t+1}}   + 1 }{n_{s:t} + O } \\
  &= \frac{ (O-1)!
  \prod\limits_{o=1}^{O}  \Sigma_{o,s+1:t+1}! }{ \left(\prod\limits_{i=1}^{O-1} \left( n_{s+1:t+1}+i \right) \right)  n_{s+1:t+1}!} \times \frac{\Sigma_{a,s:t}   + 1 }{n_{s:t} + O } \\ 
  &= \frac{ (O-1)!
  \prod\limits_{o=1}^{O}  \Sigma_{o,s+1:t+1}! }{ \left(\prod\limits_{i=1}^{O-1} \left( n_{s:t}+i \right) \right)  n_{s:t}!} \times \frac{\Sigma_{a,s:t+1}  }{n_{s:t} + O } \\
  &=  \frac{ (O-1)!
  \prod\limits_{o=1}^{O}  \Sigma_{o,s:t+1}! }{ \left(\prod\limits_{i=1}^{O-1} \left( n_{s:t}+i \right) \right)  n_{s:t}!} \times \frac{1  }{n_{s:t} + O } \\
  &=  \frac{ (O-1)!
  \prod\limits_{o=1}^{O}  \Sigma_{o,s:t+1}! }{ \left(\prod\limits_{i=1}^{O-1} \left( n_{s:t+1}+i-1 \right) \right) \left( n_{s:t+1} -1 \right) !} \times \frac{1  }{n_{s:t+1} + O -1 } \\
  &=  \frac{ (O-1)!
  \prod\limits_{o=1}^{O}  \Sigma_{o,s:t+1}! }{ \left(n_{s:t+1} + O -1 \right)  \left(\prod\limits_{i=1}^{O-1} \left( n_{s:t+1}+i-1 \right) \right) \left( n_{s:t+1} -1 \right) !} \\
  &=  \frac{ (O-1)!
  \prod\limits_{o=1}^{O}  \Sigma_{o,s:t+1}! }{   \left(\prod\limits_{i=1}^{O-1} \left( n_{s:t+1}+i \right) \right) \left( n_{s:t+1}\right) !}
\end{align*}
\end{myproof}

Notice that the cumulative loss $\hat{L}_{s:t}$ can be written as follows:

\begin{align*}
	\hat{L}_{s:t} &= \log \left( \left(n_{s:t}+O-1 \right)! \right) - \sum_{o=1}^{O} \log \left( \Sigma_{o,s:t}! \right) - \log (O-1)! \\
        &= \sum_{i=1}^{O-1} \log \left( n_{s:t}+i  \right) + \log \left( n_{s:t}! \right) - \sum_{o=1}^{O} \log \left( \Sigma_{o,s:t}! \right) - \log (O-1)!
	\end{align*}

where $n!$ denotes the factorial of $n$ such that:

\begin{align*}
    n! &= n\times (n-1)\times (n-2)\times ... \times 1
\end{align*}

Then, using the following Stirling formula:

	\begin{align*}
\forall \ n\geq 1 \quad	\sqrt{2\pi n} \left(\frac{n}{e} \right)^n  \leq n!  \leq \sqrt{2\pi n} \left(\frac{n}{e} \right)^n \exp \left(\frac{1}{12} \right),
	\end{align*}

we get the upper bound and the lower bound of the quantity $\frac{n!}{n_1! n_2! .... n_O!}$:

\begin{align}
 \frac{n^n}{n_1^{n_1}  n_2^{n_2} ... n_O^{n_O}} \times \frac{\exp(b_1) }{n^{\frac{o-1}{2}}}  \leq  \frac{n!}{n_1! n_2! .... n_O!} \leq \frac{n^n}{n_1^{n_1}  n_2^{n_2} ... n_O^{n_O}} \label{Eq:CombStirling}
\end{align}

with $\sum_{i=1}^{O} n_i = n, \ n_i \geq 0 \ \forall i \in \left\lbrace 1, ..., O \right\rbrace $ and  $b_1 = -\frac{O}{12}- \frac{O-1}{2} \log \left(2\pi \right) + \frac{O}{2} \log O$

\subsection{Upper bound of the cumulative loss for stationary observations}

Before deriving the upper bound on the cumulative loss, one should notice that:

	\begin{align}
	\Sigma_{s:t} \log \Sigma_{s:t} + \bar{\Sigma}_{s:t} \log \bar{\Sigma}_{s:t} =  \Sigma_{s:t} \log \theta + \bar{\Sigma}_{s:t} \log  \bar{\theta} + n_{s:t}\log n_{s:t}  +  n_{s:t} \KL{\frac{\Sigma_{s:t}}{n_{s:t}}}{\theta}.
	\end{align}

	\begin{align}
	\sum_{o=1}^{O} \Phi \left( \Sigma_{o,s:t} \right) = &\sum_{o=1}^O  \Sigma_{o,s:t} \log \Sigma_{o,s:t}  = \sum_{o=1}^O  \Sigma_{o,s:t} \log \theta_o + n_{s:t}\log n_{s:t}  + n_{s:t} \infdiv*{\frac{\Sigma_{1,s:t}}{n_{s:t}},...,\frac{\Sigma_{O,s:t}}{n_{s:t}}}{\theta_1,...,\theta_O} \\
\sum_{o=1}^{O} \Phi \left( \Sigma_{o,s:t} \right)= & \sum_{o=1}^O  \Sigma_{o,s:t} \log \Sigma_{o,s:t}  = \sum_{o=1}^O  \Sigma_{o,s:t} \log \theta_o + n_{s:t}\log n_{s:t}  +  n_{s:t}\infdiv*{\hat{\mu}_{1,s:t}, ..., \hat{\mu}_{O,s:t}}{\theta_1,...,\theta_O} \\
\sum_{o=1}^{O} \Phi \left( \Sigma_{o,s:t} \right)= & \sum_{o=1}^O  \Sigma_{o,s:t} \log \Sigma_{o,s:t}  = \sum_{o=1}^O  \Sigma_{o,s:t} \log \theta_o + \Phi \left( n_{s:t} \right)  + n_{s:t} \infdiv*{\hat{\mu}_{1,s:t}, ..., \hat{\mu}_{O,s:t}}{\theta_1,...,\theta_O}
\label{Eq11}
 \end{align}

Then, the upper bound of the cumulative loss takes the following form:
 
	\begin{align}
	\hat{L}_{s:t} & \overset{(a)}{\leq}
  \Phi \left( n_{s:t} \right) - \sum_{o=1}^{O} \Phi \left( \Sigma_{o,s:t} \right) + \sum_{i=1}^{O-1} \log \left( n_{s:t}+i  \right) - \log (O-1)! \label{Eq::FirstUpperBound} \\
 & \overset{(b)}{\leq} \sum_{i=1}^{O-1} \log \left( n_{s:t}+i  \right)  -  \sum_{o=1}^O  \Sigma_{o,s:t} \log \theta_o   -  n_{s:t}\infdiv*{\hat{\mu}_{1,s:t}, ..., \hat{\mu}_{O,s:t}}{\theta_1,...,\theta_O} - \log (O-1)! \nn \\
 & \overset{(c)}{\leq} \sum_{i=1}^{O-1} \log \left( n_{s:t}+i  \right)  -  \sum_{o=1}^O  \Sigma_{o,s:t} \log \theta_o - \log (O-1)!   \label{Eq::UpperBound}
	\end{align}

where:
\begin{itemize}
\setlength\itemsep{0.25em}
    \item (a) holds using the left side of Equation (\ref{Eq:CombStirling}) for $n = n_{s:t}$ and $a_o = \Sigma_{o,s:t} \  \ \forall o \in \left\lbrace 1,..., O \right\rbrace$.
    \item (b) holds thanks to the statement of Equation (\ref{Eq11}).
    \item (c) holds thanks to the fact that the Kullback Leibler divergence is always positive (i.e. $\infdiv*{\bullet}{\bullet} \geq 0$).
\end{itemize}

\subsection{Lower bound of the cumulative loss for stationary observations}

The lower bound of the cumulative loss is taking the following form:

	\begin{align}
	\hat{L}_{s:t} & \overset{(a)}{\geq}
  \Phi \left( n_{s:t} \right) - \sum_{o=1}^{O} \Phi \left( \Sigma_{o,s:t} \right) + \sum_{i=1}^{O-1} \log \left( n_{s:t}+i  \right) - \frac{O-1}{2} \log n_{s:t} + b_1 - \log (O-1)! \label{Eq::FirstLowerBound} \\
 & \overset{(b)}{\geq} \sum_{i=1}^{O-1} \log \left( n_{s:t}+i  \right)  - \sum_{o=1}^O  \Sigma_{o,s:t} \log \theta_o  -  n_{s:t}\infdiv*{\hat{\mu}_{1,s:t}, ..., \hat{\mu}_{O,s:t}}{\theta_1,...,\theta_O} - \frac{O-1}{2} \log n_{s:t} + b_1 - \log (O-1)! \label{Eq::LowerBound}  
 \end{align}

where:
\begin{itemize}
\setlength\itemsep{0.25em}
    \item (a) holds using the left side of Equation (\ref{Eq:CombStirling}) for $n = n_{s:t}$ and $n_o = \Sigma_{o,s:t} \ \ \forall o \in \left\lbrace 1,..., O \right\rbrace$
    \item (b) holds thanks to the statement of Equation (\ref{Eq11}).
\end{itemize}

\paragraph{Useful lemmas to derive the false alarm rate and detection delay}

\begin{lemma}[Time uniform $\infdiv*{\bullet}{\bullet}$ concentration]
Let: $\boldsymbol{\theta}  = \left(\theta_1,..., \theta_O \right)$ denotes the vector of the generative parameters for the Multinomial distribution $\multi{\theta_1,..., \theta_O}$.
$\forall o \in \left\lbrace 1,..., O \right\rbrace$, let $\hat{\mu}_{o,t}$ denotes the empirical frequency of observing the realization $o \in \left\lbrace 1,..., O \right\rbrace$ in the sequence $\left(x_1,...,x_t\right) \sim \multi{\theta_1,..., \theta_O}^{\otimes t}$, then for all $\left(\delta, \alpha \right) \in \OpSeg{0,1}\times \OpSeg{1, \infty}$ we have:
\begin{align*}
\Pr{\underbrace{\forall t \in \mathbb{N}^\star: \infdiv*{\hat{\mu}_{1,t}, ..., \hat{\mu}_{O,t}}{\theta_1,..., \theta_O} <  \frac{ 
\alpha  }{t}\log \frac{\log(\alpha t)\log(t)}{\log^2(\alpha)\delta}}_{E^{(1)}_{\btheta,\delta,\alpha}}}{\btheta}
\geq  1-\delta  
\end{align*}
\label{lemma::B1}
\end{lemma}

\begin{lemma}[Doubly-time uniform $\infdiv*{\bullet}{\bullet}$ concentration]
Let: $\boldsymbol{\theta}  = \left(\theta_1,..., \theta_O \right)$ denotes the vector of the generative parameters for the Multinomial distribution $\multi{\theta_1,..., \theta_O}$.

$\forall o \in \left\lbrace 1,..., O \right\rbrace$, let $\hat{\mu}_{o,s:t}$ denotes the empirical frequency of observing $o$ in the sequence $\left(x_s,...,x_t\right) \sim \multi{\theta_1,..., \theta_O}^{\otimes n_{s:t}}$, then for all $\left(\delta, \alpha \right) \in \OpSeg{0,1}\times \OpSeg{1, \infty}$ we have:
\vspace{-0.5em}
\begin{align*}
&\Pr{ \underbrace{\forall t \in \mathbb{N}^\star, \forall  s \in \segOL{r,t} : \infdiv*{\hat{\mu}_{1,s:t}, ..., \hat{\mu}_{O,s:t}}{\theta_1,..., \theta_O} <  \frac{ 
\alpha}{n_{s:t}} \times  \log \frac{n_{r:t}\log^2(n_{r:t})\log(\left(\alpha +1 \right) n_{s:t})}{\log(2)\log^2(\alpha)\delta}}_{E^{(2)}_{\btheta,\delta,\alpha}}}{\btheta}
\geq  1-\delta  
\end{align*}
\label{lemma::B2}
\end{lemma}

\begin{lemma}[Doubly-time uniform concentration]
	Let: $\left(x_{r},...x_t\right) \in \left\lbrace 1,..., O \right\rbrace^{n_{r:t}}$  be a sequence of independent  random variables sampled from a Multinomial distribution whose generative parameter can be chosen arbitrarily and $\hat{\mu}_{o,i:j}$ the empirical frequency of observing $o$ in the sequence $\left(x_i,...,x_j\right)$. Then, for all $\left(r,\delta\right) \in \mathbb{N}^\star \times \left(0,1 \right)$, we get the following control:
	\begin{align*}
	& \pr{ \exists\ t > r,  s \in \left[r, t\right): \abs{ \hat{\mu}_{o,r:s-1}-\hat{\mu}_{o,s:t}- \esp{\hat{\mu}_{o,r:s-1}-\hat{\mu}_{o,s:t}}} \geq  \pC^\prime_{r,s,t,\delta}} \leq \delta, \\
  &\pC^\prime_{r,s,t,\delta} = \frac{\sqrt{2}}{2} \Bigg( \sqrt{\frac{1+\frac{1}{n_{r:s-1}}}{n_{r:s-1}} \log \left(\frac{2\sqrt{n_{r:s}}}{\delta}\right)}  +\sqrt{\frac{1+\frac{1}{n_{s:t}}}{n_{s:t}} \log \left(\frac{2n_{r:t} \sqrt{n_{s:t}+1} \log^{2}\left(n_{r:t}\right)}{\log(2) \delta}\right)}\Bigg).
	\end{align*}
	\label{lemma::B3}
\end{lemma}

The proof of lemmas \ref{lemma::B1}, \ref{lemma::B2}, and \ref{lemma::B3} is beyond the scope of this manuscript, we refer the interested reader to section \textbf{3.4} of \cite{maillard2019mathematics}.

\section{Derivation of the false alarm rate}

\begin{myproof}{of Theorem \ref{Theorem:FalseAlarmRate}}

Let $ \boldsymbol{\theta}  = \left(\theta_1,..., \theta_O \right)$ denotes the vector of the generative parameters for the Multinomial distribution $\multi{\theta_1,..., \theta_O}^{\otimes n_{r:t}}$.

Assume that: $ \forall t \in \segOR{r, c_\ell} \ \left(x_r, ..., x_t\right) \sim \multi{\theta_1,..., \theta_O}^{\otimes n_{r:t}}$. The proof follows three  main steps:

	Let us build a suitable value of $\eta_{r,s,t}$ in order to ensure the control of the false alarm on the period $\left[r, c_\ell \right)$.
	To this end, let us control the event: $ \left\lbrace \exists t>r,  \restart{x_r,...,x_t} = 1 \right\rbrace$ which is equivalent to the event $ \left\lbrace \exists  t>r, \ s \in \left(r,t \right]: \omega_{r,s,t} \geq \omega_{r,r,t} \right\rbrace$.

\paragraph{Step 1: Equivalent events.}
	First, notice that:
	\begin{align}
	\event{ \exists  t>r, \ s \in \left(r,t \right]: \omega_{r,s,t} \geq \omega_{r,r,t} } & \equi \event{\exists  t>r, \ s \in \left(r,t \right]: \quad \log \omega_{r,s,t} \geq \log \omega_{r,r,t}}. \nn \\
	 &  \overset{(a)}{\equi} \event{ \exists  t>r, \ s \in \left(r,t \right]:- \log \eta_{r,s,t} \leq  
	\hat{L}_{r:t} - \hat{L}_{s:t} - \hat{L}_{r:s-1} } \label{Eq::Equiv}  
	\end{align}
	
	where (a) comes directly from the definition of the forecaster weights $\omega_{r,s,t}$ stated in  Equation (\ref{eq::weight}) .

	\paragraph{Step 2: Using the cumulative loss controls.}
	Then, note that $\forall \delta \in \OpSeg{0,1}, \forall \alpha > 1$ we have:
	\begin{align}
	&\mathbb{P}_{\btheta}\Big\{ \exists  t>r, \ s \in \left(r,t \right]: \omega_{r,s,t} \geq \omega_{r,r,t} \Big\} \overset{(a)}{=} \mathbb{P}_{\btheta}\Big\{ \exists  t>r, \ s \in \left(r,t \right]: \log \omega_{r,s,t} \geq \log \omega_{r,r,t} \Big\} \nn \\
	& \overset{(b)}{=} \mathbb{P}_{\btheta} \Big\{ \exists  t>r, \ s \in \left(r,t \right]: \quad - \log \eta_{r,s,t} \leq  
	\hat{L}_{r:t} - \hat{L}_{r:s-1} - \hat{L}_{s:t}  \Big\} \nn  \\
& \overset{(c)}{\leq} \mathbb{P}_{\btheta} \Big\{ \exists  t>r, \ s \in \left(r,t \right]:  - \log \eta_{r,s,t} \leq  
	\sum_{i=1}^{O-1} \log \left( n_{r:t}+i  \right)
 - \sum_{i=1}^{O-1} \log \left( n_{r:s-1}+i  \right) - \sum_{i=1}^{O-1} \log \left( n_{s:t}+i  \right) \nn
   \\
	& \phantom{\mathbb{P}_{\btheta} \Big\{ \exists  t>r, \ s \in \left(r,t \right]:  - \log \eta_{r,s,t} \leq } 
 + \frac{O-1}{2} \log n_{r:s-1} + \frac{O-1}{2} \log n_{s:t} - 2b_1 + n_{s:t}\infdiv*{\hat{\mu}_{1,s:t}, ..., \hat{\mu}_{O,s:t}}{\theta_1,...,\theta_O} \nn \\
 & \phantom{\mathbb{P}_{\btheta} \Big\{ \exists  t>r, \ s \in \left(r,t \right]:  - \log \eta_{r,s,t} \leq }
+ n_{r:s-1}\infdiv*{\hat{\mu}_{1,r:s-1}, ..., \hat{\mu}_{O,r:s-1}}{\theta_1,...,\theta_O} + \log (O-1)!
 \Bigg\} \nn \\
& \overset{(d)}{\leq} \mathbb{P}_{\btheta} \Big\{ \exists  t>r, \ s \in \left(r,t \right]:  - \log \eta_{r,s,t} \leq  
	\sum_{i=1}^{O-1} \log \frac{n_{r:t}+i}{\left( n_{r:s-1}+i  \right)\left( n_{s:t}+i  \right)}  + \frac{O-1}{2} \log \left( n_{r:s-1} n_{s:t}\right) - 2b_1  \nn 
   \\
	& \phantom{\mathbb{P}_{\btheta} \Big\{ \exists  t>r,   } 
 + n_{s:t}\infdiv*{\hat{\mu}_{1,s:t}, ..., \hat{\mu}_{O,s:t}}{\theta_1,...,\theta_O} 
+ n_{r:s-1}\infdiv*{\hat{\mu}_{1,r:s-1}, ..., \hat{\mu}_{O,r:s-1}}{\theta_1,...,\theta_O} + \log (O-1)!
 \Bigg\} \nn \\
 & \overset{(e)}{\leq} \frac{\delta}{2} + \mathbb{P}_{\btheta} \Big\{ \exists  t>r, \ s \in \left(r,t \right]:  - \log \eta_{r,s,t} \leq  
	\sum_{i=1}^{O-1} \log \frac{n_{r:t}+i}{\left( n_{r:s-1}+i  \right)\left( n_{s:t}+i  \right)}  + \frac{O-1}{2} \log \left( n_{r:s-1} n_{s:t}\right) - 2b_1 \nn
   \\
	& \phantom{\mathbb{P}_{\btheta}  } 
 + n_{s:t}\infdiv*{\hat{\mu}_{1,s:t}, ..., \hat{\mu}_{O,s:t}}{\theta_1,...,\theta_O} 
+ n_{r:s-1}\infdiv*{\hat{\mu}_{1,r:s-1}, ..., \hat{\mu}_{O,r:s-1}}{\theta_1,...,\theta_O} + \log (O-1)!
\bigcap  E^{(1)}_{\btheta,\delta/2,\alpha} \Bigg\} \nn \\
 & \overset{(f)}{\leq} \frac{\delta}{2} + \mathbb{P}_{\btheta} \Big\{ \exists  t>r, \ s \in \left(r,t \right]:  - \log \eta_{r,s,t} \leq  
	\sum_{i=1}^{O-1} \log \frac{n_{r:t}+i}{\left( n_{r:s-1}+i  \right)\left( n_{s:t}+i  \right)}  + \frac{O-1}{2} \log \left( n_{r:s-1} n_{s:t}\right) - 2b_1 \nn
   \\
	& \phantom{\mathbb{P}_{\btheta} \Big\{ \exists  t>r, } 
 + n_{s:t}\infdiv*{\hat{\mu}_{1,s:t}, ..., \hat{\mu}_{O,s:t}}{\theta_1,...,\theta_O} + \alpha\log \frac{2\log(\alpha n_{r:s-1})\log(n_{r:s-1})}{\log^2(\alpha)\delta}
+ \log (O-1)! \Bigg\} \nn \\
 & \overset{(g)}{\leq} \delta + \mathbb{P}_{\btheta} \Big\{ \exists  t>r, \ s \in \left(r,t \right]:  - \log \eta_{r,s,t} \leq  
	\sum_{i=1}^{O-1} \log \frac{n_{r:t}+i}{\left( n_{r:s-1}+i  \right)\left( n_{s:t}+i  \right)}  + \frac{O-1}{2} \log \left( n_{r:s-1} n_{s:t}\right) - 2b_1 \nn 
   \\
	& \phantom{\mathbb{P}_{\btheta} \Big\{ \exists  t>r,  } 
 + n_{s:t}\infdiv*{\hat{\mu}_{1,s:t}, ..., \hat{\mu}_{O,s:t}}{\theta_1,...,\theta_O} + \alpha\log \frac{2\log(\alpha n_{r:s-1})\log(n_{r:s-1})}{\log^2(\alpha)\delta} + \log (O-1)!
 \bigcap  E^{(2)}_{\btheta,\delta/2,\alpha}  \Bigg\} \nn \\
  & \overset{(h)}{\leq} \delta + \mathbb{P}_{\btheta} \Big\{ \exists  t>r, \ s \in \left(r,t \right]:  - \log \eta_{r,s,t} \leq  
	\sum_{i=1}^{O-1} \log \frac{n_{r:t}+i}{\left( n_{r:s-1}+i  \right)\left( n_{s:t}+i  \right)}  + \frac{O-1}{2} \log \left( n_{r:s-1} n_{s:t}\right) - 2b_1 \nn
   \\
	& \phantom{\mathbb{P}_{\btheta} \Big\{ \exists  t>r,    } 
  + \alpha\log \frac{2n_{r:t}\log^2(n_{r:t})\log(\alpha n_{s:t})\log(n_{s:t})}{\log(2)\log^2(\alpha)\delta} + \alpha\log \frac{2\log(\alpha n_{r:s-1})\log(n_{r:s-1})}{\log^2(\alpha)\delta} + \log (O-1)! \Bigg\} \nn
\\
  & \overset{(i)}{\leq} \delta + \mathbb{P}_{\btheta} \Big\{ \exists  t>r, \ s \in \left(r,t \right]:  - \log \eta_{r,s,t} \leq  
	 \log \left( \prod_{i=1}^{O-1} \frac{n_{r:t}+i}{\left( n_{r:s-1}+i  \right)\left( n_{s:t}+i  \right)} \times \left( n_{r:s-1} n_{s:t} \right)^{\frac{O-1}{2}} \right) - 2b_1 \nn 
   \\
	& \phantom{\mathbb{P}_{\btheta} \Big\{ \exists  t>r, \ s \in \left(r,t \right]:   } 
  + \alpha\log \frac{2n_{r:t}\log^2(n_{r:t})\log(\alpha n_{s:t})\log(n_{s:t})}{\log(2)\log^2(\alpha)\delta} + \alpha\log \frac{2\log(\alpha n_{r:s-1})\log(n_{r:s-1})}{\log^2(\alpha)\delta} + \log (O-1)! \Bigg\} \label{eq::CondFalseAlarm}
	\end{align}

where:
\begin{itemize}
\setlength\itemsep{0.25em}
    \item (a) holds by using the monotonic behavior of the logarithm function. 
    \item (b) holds thanks to Equation (\ref{eq::weight}).
    \item (c) holds thanks to the use of the lower bound of the cumulative loss in Equation (\ref{Eq::LowerBound}) and the upper bound of the cumulative loss in Equation (\ref{Eq::UpperBound}).
    \item (d) holds by using basic properties of the logarithm function.
    \item (e) holds by using the property that: $\pr{A}\leq \pr{\neg B} + \pr{A \cap B}$ where $B = E^{(1)}_{\btheta,\delta/2,\alpha}$.
    \item (f) holds thanks to the statement of Lemma \ref{lemma::B1}.
    \item (g) holds by using the property that: $\pr{A}\leq \pr{\neg B} + \pr{A \cap B}$ where $B = E^{(2)}_{\btheta,\delta/2,\alpha}$.
    \item (h) holds thanks to the statement of Lemma \ref{lemma::B2}.
    \item (i) holds by using the monotonic behavior of the logarithm function.
\end{itemize}

\paragraph{Step 3: Sufficient condition on $\eta_{r,s,t}$}

Based on Equation (\ref{eq::CondFalseAlarm}), we derive a sufficient condition on $\eta_{r,s,t}$ to guarantee the false alarm control:

\begin{align*}
	\eta_{r,s,t} &< \left( \prod_{i=1}^{O-1} \frac{\left( n_{r:s-1}+i  \right)\left( n_{s:t}+i  \right)}{n_{r:t}+i} \right) \times  \frac{\exp{(2b_1)}}{\left( n_{r:s-1} n_{s:t} \right)^{\frac{O-1}{2}} \times (O-1)! } \\
 &\times \left(\frac{\log^2(\alpha)\delta}{2\log(\alpha n_{r:s-1})\log(n_{r:s-1})} \times \frac{\log(2)\log^2(\alpha)\delta}{2n_{r:t}\log^2(n_{r:t})\log(\alpha n_{s:t})\log(n_{s:t})}\right)^\alpha \\
	& = 
\left(  \prod_{i=1}^{O-1} \frac{\left( n_{r:s-1}+i  \right)\left( n_{s:t}+i  \right)}{n_{r:t}+i} \right) \times \frac{\exp{(2b_1)}}{\left( n_{r:s-1} n_{s:t} \right)^{\frac{O-1}{2}} \times (O-1)!} 
 \times \left(\frac{\log(4\alpha)\log(2)\delta^2}{4n_{r:t}\log(\alpha n_{r:t})\log^2(n_{r:t})\log(n_{r:t})} \right)^\alpha\\
	& = \left( \prod_{i=1}^{O-1} \frac{\left( n_{r:s-1}+i  \right)\left( n_{s:t}+i  \right)}{n_{r:t}+i} \right) \times \frac{\exp{(2b_1)}}{\left( n_{r:s-1} n_{s:t} \right)^{\frac{O-1}{2}} \times (O-1)!} \times \left(\frac{\log(4\alpha+2)\delta^2}{4n_{r:t}\log(\left( \alpha+3 \right) n_{r:t})} \right)^\alpha
	\end{align*}

Then, no false alarm occurs with high probability during a stationary period $\segOR{r, c_\ell}$: 

	\begin{align*}
	\Pr{\exists\  t \in \segOR{r, c_\ell}: \restart{x_r,...,x_t} = 1 }{\btheta} \leq  \delta.
	\end{align*}

\end{myproof}

\section{Derivation of the detection delay}

\begin{myproof}{of Theorem \ref{Theorem:DetectionDelay}}
The proof follows three main steps:
	\paragraph{Step 1: Some preliminaries}
	Before building the detection delay, we need to introduce three intermediate results. 
	
	The first result is to link the quantity $\Phi\left(\Sigma_{o,s:t} \right)$ to $\Phi\left( \hat{\mu}_{o,s:t}  \right)$ such that:
	
 \begin{align}
	\forall \left(s,t\right): \quad  \sum_{o=1}^O \Phi\left(\Sigma_{o,s:t} \right)  - \Phi\left(n_{s:t}\right) = n_{s:t}  \sum_{o=1}^O \Phi\left( \hat{\mu}_{o,s:t}  \right).
 \label{eq::Prelim}
	\end{align} 

 Using the notation: $\hat{\bmu}_{a:b} = \left(\hat{\mu}_{1,a:b},..., \hat{\mu}_{O,a:b} \right) \quad \forall \ a < b $
 
 Then, observe that :

	\begin{align}
	&  n_{r:s-1} \sum_{o=1}^O \Phi\left( \hat{\mu}_{o,r:s-1}  \right) + n_{s:t} \sum_{o=1}^O \Phi\left( \hat{\mu}_{o,s:t} \right) 
	- n_{r:t} \sum_{o=1}^O \Phi\left( \hat{\mu}_{o,r:t}  \right)   = n_{r:s-1} \infdiv*{\hat{\bmu}_{r:s-1}}{\hat{\bmu}_{r:t}} + n_{s:t} \infdiv*{\hat{\bmu}_{s:t}}{\hat{\bmu}_{r:t}}.\nn  \\
 &  = n_{r:s-1} \infdiv*{\hat{\mu}_{1,r:s-1},..., \hat{\mu}_{O,r:s-1}}{\hat{\mu}_{1,r:t},...,\hat{\mu}_{O,r:t}} + n_{s:t} \infdiv*{\hat{\mu}_{1,s:t},...,\hat{\mu}_{O,s:t}}{\hat{\mu}_{1,r:t},..., \hat{\mu}_{O,r:t}}  \label{EqKL}
	\end{align}

	Then, observe that:
	\begin{align}
\forall o \in \left\lbrace1,...,O \right\rbrace	\quad  n_{r:s-1} \left( \hat{\mu}_{o,r:s-1} -\hat{\mu}_{o,r:t}\right)^2 +n_{s:t} \left( \hat{\mu}_{o,s:t} -\hat{\mu}_{o,r:t} \right)^2 = \frac{n_{r:s-1}n_{s:t}}{n_{r:t}} \left( \hat{\mu}_{o,r:s-1} -\hat{\mu}_{o,s:t} \right)^2. \label{Eq_Rec}
	\end{align}
	
	Then, we will also need a useful notation as $f_{r,s,t}$:
	\begin{align}
	f_{r,s,t} &= \sum_{i=1}^{O-1} \log \left( n_{r:s-1}+i  \right) + \sum_{i=1}^{O-1} \log \left( \frac{n_{s:t}+i}{n_{r:t}+i}  \right) - \frac{O-1}{2} \log \left( \frac{n_{s:t}}{n_{r:t}} \right) - \log(O-1)! \nn 
	\end{align}
	
	Finally, following Lemma \ref{lemma::B3}, the control of the quantity $\abs{\hat{\mu}_{o,r:s-1}-\hat{\mu}_{o,s:t}}$ takes the following form: 
	
	\begin{align}
	\Pr{\underbrace{\forall\  s \in \left[r:t\right) \quad
	\abs{\hat{\mu}_{o,r:s-1}-\hat{\mu}_{o,s:t}} \geq \Delta_{o,r,s,t} - \pC^\prime_{r,s,t,\delta}}_{E^{(3)}_{o,r,t,\delta}}}{\btheta^{(1)},\btheta^{(2)}} \geq 1-\delta,
	\label{LapCor}
	\end{align}

We define $\btheta^{(1)}$ and $\btheta^{(2)}$ to be the pre and post state-transition kernels over the set of actions $\sO$ for change-point $c_{\ell+1}$ and $\Delta_o$ to be the per state variation $\Delta_o = \abs{\theta^{(1)}_o -  \theta^{(2)}_o}$. Then we write the relative gap $\Delta_{r,s,t}$ as follows
	\begin{align}
 \forall o \in \left\lbrace 1,..,O \right\rbrace \ 
	\Delta_{o,r,s,t} = \abs{\Esp{\hat{\mu}_{o,r:s-1}-\hat{\mu}_{o,s:t}}{\btheta^{(1)},\btheta^{(2)}}} = 
	\begin{cases}
	\frac{n_{c_\ell:t}}{ n_{s:t}} \abs{\theta^{(1)}_o- \theta^{(2)}_o} = \frac{n_{c_\ell:t}}{ n_{s:t}} \Delta_o & \text{if }  s < c_\ell\leq t, \\
	\frac{ n_{r:c_\ell-1}}{ n_{r:s-1}}\abs{\theta^{(1)}_o- \theta^{(2)}_o} = \frac{ n_{r:c_\ell-1}}{ n_{r:s-1}} \Delta_o & \text{if }  c_\ell\leq s \leq t.
	\end{cases}
	\label{Eq::Delta}
	\end{align}

\paragraph{Pinsker inequality for multinomial distributions}
 \begin{align}
     \infdiv*{\hat{\mu}_{1,s:t},...,\hat{\mu}_{O,s:t}}{\hat{\mu}_{1,r:t},..., \hat{\mu}_{O,r:t}} 
     \geq \frac{1}{2} \left( \sum_{o=1}^{O} \abs{\hat{\mu}_{o,s:t} - \hat{\mu}_{o,r:t}} \right)^2
     \label{eq::Pinsker}
 \end{align}

	\paragraph{Step 2: Building the sufficient conditions for detecting the change-point $c_\ell$}
 Let: $\btheta^{(1)}  = \left(\theta^{(1)}_1,..., \theta^{(1)}_O \right) \in \left[0,1\right]^O$ and
$\btheta^{(2)}  = \left(\theta^{(2)}_1,..., \theta^{(2)}_O \right)$. 
	First, assume that: $x_r,...,x_{c_\ell-1}\sim \multi{\theta^{(1)}_1,..., \theta^{(1)}_O}$ and $x_{c_\ell},...,x_{t}\sim \multi{\theta^{(2)}_1,..., \theta^{(2)}_O}$. 
	Then, to build the detection delay, we need to prove that at some instant after $c_\ell$ the restart criterion $\restart{x_r,...,x_t}$ is activated. In other words, we need to build the following guarantee:
	
	\begin{align*}
	\Pr{ \exists t > c_\ell: \restart{x_r,...,x_t} = 1 }{\btheta^{(1)},\btheta^{(2)}} > 1-\delta.
	\end{align*}
	
	Notice that:
	
	\begin{align*}
	&\event{ \forall\  t > c_\ell: \restart{x_r,...,x_t} = 0}
	\overset{(a)}{\equi}
	\event{ \forall\  t > c_\ell , \forall  s \in \segOL{r,t}: \log \omega_{r,s,t} \leq \log \omega_{r,r,t}}. \\
	& \overset{(b)}{\equi} \event{\forall\  t > c_\ell , \forall  s \in \segOL{r,t}: \log \eta_{r,s,t} \leq  
		\hat{L}_{r:s-1} + \hat{L}_{s:t}-\hat{L}_{r:t} }. \\
	& \overset{(c)}{\imp} \Event{ \forall\  t > c_\ell , \forall  s \in \segOL{r,t}: \log \eta_{r,s,t} \leq  f_{r,s,t} - 
		\sum_{o=1}^O \Phi\left(\Sigma_{o,r:s-1} \right) + \Phi\left(n_{r:s-1}\right) - \sum_{o=1}^O \Phi\left(\Sigma_{s:t} \right) + \Phi\left(n_{s:t}\right)  \\
&\phantom{ \overset{(c)}{\imp} \Event{ \forall\  t > c_\ell , \forall  s \in \segOL{r,t}: \log \eta_{r,s,t} \leq  f_{r,s,t} - 
		\sum_{o=1}^O \Phi\left(\Sigma_{o,r:s-1} \right) + \Phi\left(n_{r:s-1}\right)}}  + \sum_{o=1}^O \Phi\left(\Sigma_{r:t} \right)  - \Phi\left(n_{r:t}\right)}
	 . \\  
  & \overset{(d)}{\imp} \Event{ \forall\  t > c_\ell , \forall  s \in \segOL{r,t}: \log \eta_{r,s,t} \leq  f_{r,s,t} - 
		n_{r:s-1}  \sum_{o=1}^O \Phi\left( \hat{\mu}_{o,r:s-1}  \right) - n_{s:t}  \sum_{o=1}^O \Phi\left( \hat{\mu}_{o,s:t}  \right) + n_{r:t}  \sum_{o=1}^O \Phi\left( \hat{\mu}_{o,r:t}  \right) }
  \\
  & \overset{(e)}{\imp} \Event{ \forall\  t > c_\ell , \forall  s \in \segOL{r,t}: \log \eta_{r,s,t} \leq  f_{r,s,t} - 
		n_{r:s-1} \infdiv*{\hat{\mu}_{1,r:s-1},..., \hat{\mu}_{O,r:s-1}}{\hat{\mu}_{1,r:t},...,\hat{\mu}_{O,r:t}} \\
  &\phantom{\forall\  t > c_\ell , \forall  s \in \segOL{r,t}: \log \eta_{r,s,t} \leq  f_{r,s,t} -}
  - n_{s:t} \infdiv*{\hat{\mu}_{1,s:t},...,\hat{\mu}_{O,s:t}}{\hat{\mu}_{1,r:t},..., \hat{\mu}_{O,r:t}} } \\
& \overset{(f)}{\imp} \Eventt{ \forall\  t > c_\ell , \forall  s \in \segOL{r,t}: \log \eta_{r,s,t} \leq  f_{r,s,t} - 
		\frac{n_{r:s-1}}{2} \left( \sum_{o=1}^{O} \abs{\hat{\mu}_{o,r:s-1} - \hat{\mu}_{o,r:t}} \right)^2 
  -\frac{n_{s:t}}{2} \left( \sum_{o=1}^{O} \abs{\hat{\mu}_{o,s:t} - \hat{\mu}_{o,r:t}} \right)^2 } \\
& \overset{(g)}{\imp} \Eventt{ \forall\  t > c_\ell , \forall  s \in \segOL{r,t}: \log \eta_{r,s,t} \leq  f_{r,s,t} - 
		\frac{n_{r:s-1}}{2}  \sum_{o=1}^{O} \left(\hat{\mu}_{o,r:s-1} - \hat{\mu}_{o,r:t}\right)^2  
  -\frac{n_{s:t}}{2}  \sum_{o=1}^{O} \left(\hat{\mu}_{o,s:t} - \hat{\mu}_{o,r:t}\right)^2  } \\
& \equi \Eventt{ \forall\  t > c_\ell , \forall  s \in \segOL{r,t}: \log \eta_{r,s,t} \leq  f_{r,s,t} - 
		\frac{1}{2}  \sum_{o=1}^{O} \left( n_{r:s-1} \left(\hat{\mu}_{o,r:s-1} - \hat{\mu}_{o,r:t}\right)^2  
     +n_{s:t} \left(\hat{\mu}_{o,s:t} - \hat{\mu}_{o,r:t}\right)^2 \right)  }
     \\
& \overset{(h)}{\imp} \Eventt{ \forall\  t > c_\ell , \forall  s \in \segOL{r,t}: \log \eta_{r,s,t} \leq  f_{r,s,t} - 
		\frac{1}{2}  \sum_{o=1}^{O} \frac{n_{r:s-1}n_{s:t}}{n_{r:t}} \left( \hat{\mu}_{o,r:s-1} -\hat{\mu}_{o,s:t} \right)^2  }
  \\
& \equi \Eventt{ \forall\  t > c_\ell , \forall  s \in \segOL{r,t}: \frac{1}{2}  \sum_{o=1}^{O} \frac{n_{r:s-1}n_{s:t}}{n_{r:t}} \left( \hat{\mu}_{o,r:s-1} -\hat{\mu}_{o,s:t} \right)^2  \leq  f_{r,s,t} - \log \eta_{r,s,t} }
  \end{align*}

where:
\begin{itemize}
\setlength\itemsep{0.1em}
    \item (a) holds thanks to the definition of the restart procedure in Equation (\ref{eq::Restart}).
    \item (b) holds thanks to the statement of Equation (\ref{eq::weight}).
    \item (c) holds thanks to the upper bound in Equation (\ref{Eq::FirstUpperBound}) and 
    \item (d) holds thanks to the statement of Equation (\ref{eq::Prelim}).
    \item (e) holds thanks to the statement of Equation (\ref{EqKL}).
    \item (f) holds thanks to Equation (\ref{eq::Pinsker}).
    \item (g) holds thanks to the following equation: $    \left( \sum_{o=1}^{O} \abs{\hat{\mu}_{o,s:t} - \hat{\mu}_{o,r:t}} \right)^2 \geq  \sum_{o=1}^{O} \left(\hat{\mu}_{o,s:t} - \hat{\mu}_{o,r:t} \right)^2$
    \item (h) holds thanks to Equation (\ref{Eq_Rec}).
\end{itemize}  

Thus we have:

\begin{align}
    \event{ \forall\  t > c_\ell: \restart{x_r,...,x_t} = 0} \imp \Eventt{ \forall\  t > c_\ell , \forall  s \in \segOL{r,t}: \frac{1}{2}  \sum_{o=1}^{O} \frac{n_{r:s-1}n_{s:t}}{n_{r:t}} \left( \hat{\mu}_{o,r:s-1} -\hat{\mu}_{o,s:t} \right)^2  \leq  f_{r,s,t} - \log \eta_{r,s,t} }
    \label{Eq::Implication}
\end{align}

Then, by using the probability operator, we obtain:

\begin{align*}
    &\Pr{ \forall\  t > c_\ell: \restart{x_r,...,x_t} = 0}{\btheta^{(1)},\btheta^{(2)}} \\
    &\overset{(k)}{\leq}  \Pr{ \forall\  t > c_\ell , \forall  s \in \segOL{r,t}: \frac{1}{2}  \sum_{o=1}^{O} \frac{n_{r:s-1}n_{s:t}}{n_{r:t}} \left( \hat{\mu}_{o,r:s-1} -\hat{\mu}_{o,s:t} \right)^2  \leq  f_{r,s,t} - \log \eta_{r,s,t} }{\btheta^{(1)},\btheta^{(2)}} \\ 
    & \overset{(l)}{\leq} \Pr{\neg \Event{ \bigcap_{o \in \mathcal{O}} E^{(3)}_{o,r,t,\delta^\prime}}}{\btheta^{(1)},\btheta^{(2)}} \\
    &+ \Pr{ \forall\  t > c_\ell , \forall  s \in \segOL{r,t}: \frac{1}{2}  \sum_{o=1}^{O} \frac{n_{r:s-1}n_{s:t}}{n_{r:t}} \left( \hat{\mu}_{o,r:s-1} -\hat{\mu}_{o,s:t} \right)^2  \leq  f_{r,s,t} - \log \eta_{r,s,t} \bigcap \Event{ \bigcap_{o \in \mathcal{O}} E^{(3)}_{o,r,t,\delta^\prime}} }{\btheta^{(1)},\btheta^{(2)}} \\
    & \overset{(m)}{\leq} \Pr{ \bigcup_{o \in \mathcal{O}} \neg E^{(3)}_{o,r,t,\delta^\prime}}{\btheta^{(1)},\btheta^{(2)}} \\
    &+ \Pr{ \forall\  t > c_\ell , \forall  s \in \segOL{r,t}: \frac{1}{2}  \sum_{o=1}^{O} \frac{n_{r:s-1}n_{s:t}}{n_{r:t}} \left( \hat{\mu}_{o,r:s-1} -\hat{\mu}_{o,s:t} \right)^2  \leq  f_{r,s,t} - \log \eta_{r,s,t} \bigcap \Event{ \bigcap_{o \in \mathcal{O}} E^{(3)}_{o,r,t,\delta^\prime}} }{\btheta^{(1)},\btheta^{(2)}} \\
    & \overset{(n)}{\leq} \sum_{o \in \mathcal{O}} \Pr{\neg E^{(3)}_{o,r,t,\delta^\prime}}{\btheta^{(1)},\btheta^{(2)}} \\
    &+ \Pr{ \forall\  t > c_\ell , \forall  s \in \segOL{r,t}: \frac{1}{2}  \sum_{o=1}^{O} \frac{n_{r:s-1}n_{s:t}}{n_{r:t}} \left( \hat{\mu}_{o,r:s-1} -\hat{\mu}_{o,s:t} \right)^2  \leq  f_{r,s,t} - \log \eta_{r,s,t} \bigcap \Event{ \bigcap_{o \in \mathcal{O}} E^{(3)}_{o,r,t,\delta^\prime}} }{\btheta^{(1)},\btheta^{(2)}} \\
    & \overset{(o)}{\leq} O\delta^\prime + \Pr{ \forall\  t > c_\ell , \forall  s \in \segOL{r,t}: \frac{1}{2}  \sum_{o=1}^{O} \frac{n_{r:s-1}n_{s:t}}{n_{r:t}} \left( \Delta_{o,r,s,t} - \pC^\prime_{r,s,t,\delta^\prime} \right)^2  \leq  f_{r,s,t} - \log \eta_{r,s,t} }{\btheta^{(1)},\btheta^{(2)}} \\
	& \equi O\delta^\prime +  \Pr{\forall\  t > c_\ell , \forall  s \in \segOL{r,t}: \ 
	\underbrace{1- \frac{f_{r,s,t} - \log \eta_{r,s,t}}{
	\frac{n_{r,s-1}}{2} \times \sum_{o=1}^{O} \left(\Delta_{o,r,s,t} - \pC^\prime_{r,s,t,\delta^\prime}  \right)^2}}_{A_{r,s,t, \delta^\prime}} \leq \frac{n_{r:s-1}}{n_{r:t}}}{\btheta^{(1)},\btheta^{(2)}}
\end{align*}

where:
\begin{itemize}
\setlength\itemsep{0.12em}
    \item (k) holds thanks to the implication in Equation (\ref{Eq::Implication}).
    \item (l) holds by using the property that: $\pr{A}\leq \pr{\neg B} + \pr{A \cap B}$ where $B = \bigcap_{o \in \mathcal{O}} E^{(3)}_{o,r,t,\delta^\prime}$.
    \item (m) holds thanks to the fact that: $\neg \Event{ \bigcap_{o \in \mathcal{O}} E^{(3)}_{o,r,t,\delta^\prime}} =  \bigcup_{o \in \mathcal{O}} \neg E^{(3)}_{o,r,t,\delta^\prime}$.
    \item (n) holds thanks to the use of a union bound on the event $ \bigcup_{o \in \mathcal{O}} \neg E^{(3)}_{o,r,t,\delta^\prime}$.
    \item (o) holds using Equation (\ref{LapCor}).
\end{itemize}  

Then, in order to derive the detection delay, some conditions on the $A_{r,s,t, \delta^\prime}$ quantity should meet.

\paragraph{Conditions on $A_{r,s,t, \delta^\prime}$  to derive the detection delay:}
	\begin{align}
	\begin{cases}
	A_{r,s,t, \delta^\prime} > 0 & \equi \eta_{r,s,t} > \exp\left(- \frac{n_{r,s-1}}{2} \times \sum_{o=1}^{O} \left(\Delta_{o,r,s,t} - \pC^\prime_{r,s,t,\delta^\prime}  \right)^2 \right) \exp\left(f_{r,s,t} \right), \label{GammaLwBnd}
	\\
	A_{r,s,t, \delta^\prime} < 1 & \equi 
	\eta_{r,s,t} <  \exp\left( f_{r,s,t} \right)  
	\end{cases}
	\end{align}

\paragraph{Main implication for detecting the change-point.}
Notice that:

\begin{align}
	&\Event{\exists t > c_\ell, s \in \segOL{r,t}:  1+ \frac{\log \eta_{r,s,t} - f_{r,s,t}}{
			\frac{n_{r,s-1}}{2} \times \sum_{o=1}^{O} \left(\Delta_{o,r,s,t} - \pC^\prime_{r,s,t,\delta}  \right)^2} > \frac{n_{r:s-1}}{n_{r:t}} } \nn \\
   &\equi \Pr{\forall\  t > c_\ell , \forall  s \in \segOL{r,t}: \ 
	1- \frac{f_{r,s,t} - \log \eta_{r,s,t}}{
	\frac{n_{r,s-1}}{2} \times \sum_{o=1}^{O} \left(\Delta_{o,r,s,t} - \pC^\prime_{r,s,t,\delta^\prime}  \right)^2} \leq \frac{n_{r:s-1}}{n_{r:t}}}{\btheta^{(1)},\btheta^{(2)}}=0 \nn \\
   & \imp \Pr{ \forall\  t > c_\ell: \restart{x_r,...,x_t} = 0}{\btheta^{(1)},\btheta^{(2)}} \leq O \delta^\prime \equi
	\Pr{\exists t > c_\ell: \restart{x_r,...,x_t} = 1}{\btheta^{(1)},\btheta^{(2)}} > 1- O\delta^\prime. \label{Eq::ChangePointDetection}
	\end{align}		

Let $\delta = O\delta^\prime$ and $\pC_{r,s,t,\delta} = \pC^\prime_{r,s,t,\frac{\delta}{O}}$ 
 
Then, using the result of Equation (\ref{Eq::ChangePointDetection}), the change-point $c_\ell$ is detected at time $t$ (with probability at least $1-\delta$) if for some $s \in \segOL{r,t}$, we have:
	\begin{align}
	1+ \frac{\log \eta_{r,s,t} - f_{r,s,t} }{			\frac{n_{r,s-1}}{2} \times \sum_{o=1}^{O} \left(\Delta_{o,r,s,t} - \pC_{r,s,t,\delta}  \right)^2} > \frac{n_{r:s-1}}{n_{r:t}}.
	\label{Eq_00}
	\end{align}

Let $\bDelta = \left(\Delta_1,...,\Delta_O \right)$ denotes the vector of the change-point gap.
 
	\paragraph{Step 3: Non-asymptotic expression of the detection delay $\kD_{\bDelta,r,c_\ell}$}	
	
	To build the detection delay, we need to ensure the existence of $s \in \segOL{r,t}$ such that Equation (\ref{Eq_00}) is satisfied. In particular, Equation (\ref{Eq_00}) can be satisfied for $s = c_\ell$.
	By this way, a condition to detect the change-point $c_\ell$ is written as follows
	
	\begin{align}
	&		1+ \frac{\log \eta_{r,c_\ell,t} - f_{r,c_\ell,t} }{			\frac{n_{r,s-1}}{2} \times \sum_{o=1}^{O} \left(\Delta_{o} - \pC_{r,s,t,\delta}  \right)^2} > \frac{n_{r:c_\ell-1}}{n_{r:t}}.
	\label{Eq20}
	\end{align}

	To build the delay, we should introduce the following variable: $d = t-c_\ell+1 = n_{c_\ell:t} \in \mathbb{N}^\star $.
	
	Thus from Equation (\ref{Eq20}), we obtain:
	
	\begin{align*}
	& \Eventt{1+ \frac{\log \eta_{r,c_\ell,d+c_\ell-1} - f_{r,c_\ell,d+c_\ell-1}}{\frac{n_{r,s-1}}{2} \times \sum_{o=1}^{O} \left(\Delta_{o} - \pC_{r,s,t,\delta}  \right)^2} > \frac{n_{r:c_\ell-1}}{n_{r:c_\ell-1} + d}}.
  \\
	& \equi \Eventt{d > \frac{2}{\sum_{o=1}^{O}\left(\Delta_o - \pC_{r,c_\ell, d+ c_\ell-1 ,\delta}  \right)^2}\times \frac{-\log \eta_{r,c_\ell,d+c_\ell-1} + f_{r,c_\ell,d+c_\ell-1}}{1+\frac{2\left(\log \eta_{r,c_\ell,d+c_\ell-1} - f_{r,c_\ell,d+c_\ell-1}\right)}{n_{r,c_\ell-1}\times   \sum_{o=1}^{O}\left(\Delta_o - \pC_{r,c_\ell, d+ c_\ell-1 ,\delta}  \right)^2}}}.
	\end{align*}

	Finally, the change-point $c_\ell$ is detected (with a probability at least $1-\delta$) with a delay not exceeding $\kD_{\bDelta,r,c_\ell}$, such that:
	\begin{align*}
	\kD_{\bDelta,r,c_\ell} =  \min \Eventt{d \in \mathbb{N}^\star: d > \frac{2}{\sum_{o=1}^{O}\left(\Delta_o - \pC_{r,c_\ell, d+ c_\ell-1 ,\delta}  \right)^2}\times \frac{-\log \eta_{r,c_\ell,d+c_\ell-1} + f_{r,c_\ell,d+c_\ell-1}}{1+\frac{2\left(\log \eta_{r,c_\ell,d+c_\ell-1} - f_{r,c_\ell,d+c_\ell-1}\right)}{n_{r,c_\ell-1}\times  \sum_{o=1}^{O}\left(\Delta_o - \pC_{r,c_\ell, d+ c_\ell-1 ,\delta}  \right)^2}}}.
	\end{align*}

\end{myproof}





\newpage

\section{R-BOCPD equipped UCRL2 Analysis}
We consider a formulation of the regret as defined in Section \ref{sec: setting}. Given the nature of the theoretical guarantees provided by the \texttt{R-BOCPD} algorithm, we adopt a decomposition with respect to the change points, allowing to analyze the \textit{switching-MDP} problem $\mathbf{M} = \{\mathbb{S}=\{M_0,..,M_{K_T-1}\}, \mathcal{C}=\{c_0,..,c_{K_T}\}\}$ as a sequence of stationary MDPs $M_\ell$ over time instances $t \in \left[c_\ell, c_{\ell+1}\right)$. This can be formulated as follows:

\begin{align*}
\mathfrak{R}\left(\mathbf{M}, \texttt{R-BOCPD-UCRL2}, o, T\right) & =\sum_{t=1}^T\left(\rho_{\mathbf{M_\ell}}^\star(t)- \esp{r_t} \right) \\
& = \sum_{\ell=0}^{K_T-1}  \sum_{t=c_\ell}^{c_{\ell+1}-1} \left(\rho_{\mathbf{M_\ell}}^\star(t)- \esp{r_t}  \right) \\
\end{align*}

where we denote by $c_\ell$ the time instance change $\ell$ happens, denote by $t=c_\ell$ the time instance starting at $c_\ell$ up to but not including $c_{\ell+1}$, i.e $t \in [c_\ell, c_\ell+1)$, and define $r_t$ to be the random reward \texttt{UCRL2} receives at time instant $t$, when starting at some initial state $o$. Again, this decomposition is only possible by using the independence of the sum of rewards/regrets in the stationary periods of the initial state as in \cite{puterman2014markov}.\\

Now, denote by ${d_\ell}$ the detection delay achieved by \texttt{R-BOCPD} in a given interval $\left[c_\ell, c_{\ell+1}\right)$. Hence, the natural decomposition into a stationary period $\left[c_\ell+d_{\ell}, c_{\ell+1}\right)$ and a detection phase $\left[c_\ell, c_\ell+d_{\ell}\right)$.

\subsection{Detection Phase $\left[c_\ell, c_\ell+d_{\ell}\right)$}
During the detection phase, we suppose the algorithm assumes the worst possible regret of 1, as $r_t$ is sampled according to some unknown distribution in $\left[0, 1\right]$. To minimize the total regret, we rely on R-BOPCD's minimal detection delay, which we write as
	\begin{align}\label{eq:delay_cl}
	\kD_{\bDelta_\ell,r,c_\ell} =  \min \Eventt{d_\ell \in \mathbb{N}^\star: d_\ell > \frac{2}{\sum_{o=1}^{O}\left(\Delta_o - \pC_{r,c_\ell, d_\ell+ c_\ell-1 ,\delta}  \right)^2}\times \frac{-\log \eta_{r,c_\ell,d_\ell+c_\ell-1} + f_{r,c_\ell,d_\ell+c_\ell-1}}{1+\frac{2\left(\log \eta_{r,c_\ell,d_\ell+c_\ell-1} - f_{r,c_\ell,d_\ell+c_\ell-1}\right)}{n_{r,c_\ell-1}\times   \textcolor{black}{\sum_{o=1}^{O}\left(\Delta_o - \pC_{r,c_\ell, d_\ell+ c_\ell-1 ,\delta}  \right)^2}}}}
	\end{align}

with 

\begin{align*}
\bDelta_\ell = \left(\Delta_{1,\ell},...,\Delta_{O,\ell} \right) \quad \text{where} \quad \Delta_{o,\ell} =  \abs{\theta^{(\ell-1)}_o - \theta^{(\ell)}_o}
\end{align*}

which holds with probability at least $1-\delta_{1_\ell}$ for $\delta_{1_\ell} \in (0, 1)$, for an input stream starting at some time instance $r < c_\ell$. This allows us to write the regret for $t \in \left[c_\ell, c_\ell+d_{\ell}\right)$ starting at some state $o_{c_\ell}$ as follows 

\begin{align*}
\mathfrak{R}\left(M_\ell, \texttt{R-BOCPD-UCRL2}, o_{c_\ell}, \left[c_\ell, c_\ell+d_{\ell}\right) \right) & =\sum_{t=c_\ell}^{c_\ell+d_{\ell}-1}\left(\rho_{\mathbf{M_\ell}}^\star(t)- \esp{r_t} \right) \\
& = (c_\ell+d_{\ell}-1 - c_\ell + 1) \cdot 1 \\
& = d_\ell \\
& = \kD_{\bDelta_\ell,c_{\ell-1}+d_{\ell-1},c_\ell}
\end{align*}

where $c_{\ell-1}+d_{\ell-1}$ corresponds to the maximally delayed restart time after change-point $c_{\ell-1}$ with probability at least $1-\delta_{1_{\ell-1}}-\delta_{2_{\ell-1}}+\delta_{1_{\ell-1}}\delta_{2_{\ell-1}}$, where $\delta_{1_{\ell-1}}$ corresponds to the probability of the \texttt{R-BOCPD} delay exceeding $\kD_{\bDelta_{\ell-1},c_{\ell-2}+d_{\ell-2},c_{\ell-1}}$ for change-point $c_{\ell-1}$ and $\delta_{2_{\ell-1}}$ corresponds to the worst-case false-alarm probability on the stationary period $\left[c_{\ell-1}+d_{\ell-1}, c_{\ell}\right)$.

\subsection{Post-Detection Phase $\left[c_\ell+d_{\ell}, c_{\ell+1}\right)$ and Episodic Regret}
Relying on the assumptions about the change-point generating process in Section \ref{sec: setting}, we now analyze the regret in the phase $\left[c_\ell+d_{\ell}, c_{\ell+1}\right)$ assuming \texttt{R-BOCPD-UCRL2} restarts \texttt{UCRL2} exactly at time $t=c_\ell + d_\ell$ with probability at least $1-\delta_{1_\ell}$ for $\delta_{1_\ell} \in (0, 1)$. Now, to perform a similar analysis to that of \cite{auer2008near}, we need to ensure that no restarts/false-alarms happen during $\left[c_\ell+d_{\ell}, c_{\ell+1}\right)$, i.e it is stationary. Given that \texttt{R-BOCPD} guarantees a bounded probability of false-alarm, we adopt a decomposition with regards to the event of restarting \texttt{UCRL2} during a stationary period. More precisely, we make use of the concentration characteristic of the stationary sum of rewards for \texttt{UCRL2} as in \cite{auer2008near} along with the $\delta$-bound guarantee in the \texttt{R-BOCPD} false-alarm probability. 

To use the concentration argument for the sum of rewards when applying \texttt{UCRL2}, we note that at time instance $t$, reward $r_{t+1}$ is only dependent on reward $r_t$ (and filtration history $(o_1, a_1, r_1, ..., o_t, a_t, r_t)$ henceforth) through an exogenous process $\mathcal{E}$. Hence $r_{t+1}$ and $r_t$ are independent given $\mathcal{E}$ for all $t$, or $r_{t+1} \indep r_t | \mathcal{E}$. This allows us to write, by virtue of Hoeffding's inequality, for $t \in \left[c_\ell+d_{\ell}, c_{\ell+1}\right)$ and $\delta_{\ell} \in (0, 1)$ 





\begin{align*}
& \mathbb{P}\left[\underbrace{\sum_{t=c_\ell+d_{\ell}}^{c_{\ell+1}-1}r_t  
\leq \sum_{o,a} N_{\ell}(o,a)\Bar{r}_{\ell}(s,a)-\sqrt{\frac{5}{8}\left(c_{\ell+1}-\left( c_{\ell}+d_{\ell} \right) \right)\log\left(\frac{8\left(c_{\ell+1}-\left( c_{\ell}+d_{\ell} \right) \right)}{\delta_{\ell}}\right)} \Bigg|\ \left(N_\ell(o,a) \right)_{o,a}, \mathcal{E}}_{E^{(4)}_{\ell}}\right] \\
& \overset{(a)}{\leq} \mathbb{P}_{\btheta}\Bigl[ E^{(4)}_{\ell}\:\Big|\: \forall t \in  \left[c_\ell+d_{\ell}+1, c_{\ell+1}\right):
 \:\restart{o_{c_\ell+d_{\ell}},...,o_t}=0\Bigr] \\
 & \hspace{0cm} + \mathbb{P}_{\btheta}\Bigl[ \exists\  t \in \left[c_\ell+d_{\ell}+1, c_{\ell+1}\right):
 \restart{o_{c_\ell+d_{\ell}},...,o_t} = 1\Bigl]\\
 & \overset{(b)} \leq \left(\frac{\delta_\ell}{8\left(c_{\ell+1}-\left( c_{\ell}+d_{\ell} \right)\right)}\right)^{5/4} + \delta_{2_\ell} \\ 
 & < \frac{\delta_\ell}{12\left(c_{\ell+1}-\left( c_{\ell}+d_{\ell} \right)\right)^\frac{5}{4}} + \delta_{2_\ell}
\end{align*}

where (a) originates from the inequality $\mathbb{P}(A) \leq \mathbb{P}(A|B) + \mathbb{P}(\neg B)$, where $A  =  E^{(4)}_{\ell}$ and $B = \Event{ \forall t \in  \left(c_\ell+d_{\ell}+1, c_{\ell+1}\right): \:\restart{o_{c_\ell+d_{\ell}},...,o_t}=0}$ and (b) originates from Hoeffding's inequality for vanilla \texttt{UCRL2} with $\overline{r}_{\ell}(o,a) = \frac{1}{N_{\ell}(o,a)}\sum\limits_{t= c_{\ell}+d_{\ell}}^{c_{\ell+1}-1} r_t \cdot \indFct{o_t=o,a_t=a}$, in addition to \texttt{R-BOCPD}'s guarantee on false-alarm rate. \\ 

Thus, we can express the post-detection regret for the $\ell^\text{th}$ change-point starting at some state $o_{c_\ell}$ as follows

\begin{align*}
    & \mathfrak{R}\left(M_\ell, \texttt{R-BOCPD-UCRL2}, o_{c_\ell}, \left[c_\ell+d_{\ell}, c_{\ell+1}\right) \right)= (c_{\ell+1}-c_\ell - d_\ell)\rho_{\mathbf{M_\ell}}^\star-\sum_{t=c_{\ell}+d_{\ell} }^{c_{\ell+1}-1} r_t \\
    & < (c_{\ell+1}-c_\ell - d_\ell) \rho_{\mathbf{M_\ell}}^\star - \sum_{o,a} N_{\ell}(o,a)\Bar{r_{\ell}}(o,a)+\sqrt{\frac{5}{8}\left(c_{\ell+1}- c_{\ell}- d_{\ell} \right)\log\left(\frac{8\left(c_{\ell+1}- c_{\ell}-d_{\ell} \right)}{\delta_\ell}\right)}
\end{align*}

with probability at least $1-\frac{\delta_\ell}{12\left(c_{\ell+1}-c_{\ell}-d_{\ell}\right)^\frac{5}{4}} -\delta_{2_\ell}$. As in \cite{auer2008near}, we adopt a decomposition over the number of episodes, which we denote by $m_\ell$ for change interval  $\left[c_\ell+d_{\ell}, c_{\ell+1}\right)$. Consequently, we can write $\sum\limits_{k=1}^{m_\ell}\nu_{k} = N_\ell(o,a)$ and $\sum\limits_{o, a} N_\ell(o,a) = c_{\ell+1}- c_{\ell}-d_{\ell}$, where $\nu_k (o, a)$ denotes the final counts of state-action pair $(o, a)$ in episode $k$. Hence, defining $\mathfrak{R}_k \left(M_\ell, \texttt{R-BOCPD-UCRL2}, o_{c_\ell}, \left[c_\ell+d_{\ell}, c_{\ell+1}\right) \right) \dfq \sum\limits_{o,a}\nu_{k}(o,a) \left(\rho_{M_\ell}^\star-\overline{r_{\ell}}(o,a) \right)$, we can write

\begin{align*}
    & \mathfrak{R} \left(M_\ell, \texttt{R-BOCPD-UCRL2}, o_{c_\ell}, \left[c_\ell+d_{\ell}, c_{\ell+1}\right) \right) \\
    & \leq \sum_{k=1}^{m_\ell} \mathfrak{R}_k \left(M_\ell, \texttt{R-BOCPD-UCRL2}, o_{c_\ell}, \left[c_\ell+d_{\ell}, c_{\ell+1}\right) \right) 
     + \sqrt{\frac{5}{8}\left(c_{\ell+1}-c_{\ell}-d_{\ell}\right)\log\left(\frac{8\left(c_{\ell+1}- c_{\ell}+d_{\ell}\right)}{\delta_\ell}\right)}
\end{align*}

with probability at least $1-\frac{\delta_\ell}{12\left(c_{\ell+1}- c_{\ell}-d_{\ell}\right)^\frac{5}{4}} -\delta_{2_\ell}$. 

Now, following the analysis of \cite{auer2008near} for vanilla \texttt{UCRL2}, we can derive the final regret bound for \texttt{R-BOCPD-UCRL2} in post-detection period $\left[c_\ell+d_{\ell}, c_{\ell+1}\right)$, for $c_{\ell+1}-c_{\ell}-d_{\ell} > 1$, as follows 
\begin{align*}
    \mathfrak{R} \left(M_\ell, \texttt{R-BOCPD-UCRL2}, o_{c_\ell}, \left[c_\ell+d_{\ell}, c_{\ell+1}\right) \right) \leq 34 D_\ell O \sqrt{A \left(c_{\ell+1}-c_{\ell}-d_{\ell} \right) \log\left(\frac{c_{\ell+1}-c_{\ell}-d_{\ell}}{\delta_\ell} \right)}
\end{align*}

which holds with probability at least $1-\frac{\delta_\ell}{4\left(c_{\ell+1}- c_{\ell}-d_{\ell}\right)^\frac{5}{4}} -\delta_{2_\ell}$, where $D_\ell$ is the diameter of MDP $M_\ell$ as defined in \ref{sec: setting}. Also note that $d_0 = \kD_{\bDelta,.,c_0} = 0$ as time instance $c_0$ defines the start of learning.


\subsection{Total Regret Bound}
Wrapping up the last two steps and summing over the change periods, we can write 
\begin{align}
& \sum\limits_{\ell=0}^{K_T-1} \mathfrak{R} \left(M_\ell, \texttt{R-BOCPD-UCRL2}, o_{c_\ell}, \left[c_\ell, c_{\ell+1}\right) \right) \nonumber \\ 
& \leq 34 O \sqrt{A} \sum\limits_{\ell=0}^{K_T-1} D_\ell \sqrt{\left(c_{\ell+1}-c_{\ell}-d_{\ell} \right) \log\left(\frac{c_{\ell+1}-c_{\ell}-d_{\ell}}{\delta_\ell} \right)} + \sum\limits_{\ell=0}^{K_T-1} \kD_{\bDelta_{\ell+1},c_{\ell}+d_{\ell},c_{\ell+1}}
\end{align}
with probability at least $1-\sum\limits_{\ell=0}^{K_T-1} \left(\frac{\delta_\ell}{4\left(c_{\ell+1}- c_{\ell}-d_{\ell}\right)^\frac{5}{4}} + \delta_{2_\ell}\right) - \sum\limits_{\ell=1}^{K_T} \left( \delta_{1_{\ell-1}} +  \delta_{2_{\ell-1}} -  \delta_{1_{\ell-1}}\delta_{2_{\ell-1}}\right)$. Now we conclude the proof by providing a bound for the latter probability. Without loss of generality, we fix our confidence parameters as $\delta_\ell = \delta_{1_\ell} = \delta_{2_\ell} \dfq \frac{\delta}{8 K_T}, \forall \ell$. Hence, we can write:

\begin{align*}
     1-\sum\limits_{\ell=0}^{K_T-1} \left(\frac{\delta_\ell}{4\left(c_{\ell+1}- c_{\ell}-d_{\ell}\right)^\frac{5}{4}} + \delta_{2_\ell}\right) - \sum\limits_{\ell=1}^{K_T} \left( \delta_{1_{\ell-1}} +  \delta_{2_{\ell-1}} -  \delta_{1_{\ell-1}}\delta_{2_{\ell-1}}\right) \\ 
     & \hspace{-130pt} > 1-\sum\limits_{\ell=0}^{K_T-1} \left(\frac{\delta_\ell}{4\left(c_{\ell+1}- c_{\ell}-d_{\ell}\right)^\frac{5}{4}} + \delta_{2_\ell}\right) - \sum\limits_{\ell=1}^{K_T} \left( \delta_{1_{\ell-1}} +  \delta_{2_{\ell-1}} \right) \\
     & \hspace{-130pt} > 1 - \frac{3 \delta}{8}  -  \frac{\delta}{4 K_T} \sum\limits_{\ell=0}^{K_T-1} \frac{1}{\left(c_{\ell+1}- c_{\ell}-d_{\ell}\right)^\frac{5}{4}} \\
     & \hspace{-130pt} > 1 - \frac{3 \delta}{8}  -  \frac{\delta}{4 K_T} \sum\limits_{\ell=0}^{K_T-1} 1 \qquad \text{as} \quad c_{\ell+1}- c_{\ell}-d_{\ell} > 1 \\
     & \hspace{-130pt} > 1 - \frac{3 \delta}{8}  -  \frac{\delta}{4} = 1 - \frac{5 \delta}{8} \\
     & \hspace{-130pt} > 1 - \delta \\
\end{align*}
Now, defining $D \dfq \max\limits_\ell D_\ell$, deriving the corresponding regret boils down to 
\begin{align}
    \mathfrak{R} \left(\mathbf{M}, \texttt{R-BOCPD-UCRL2}, o_{c_0}, T\right) & = \sum\limits_{\ell=0}^{K_T-1} \mathfrak{R} \left(M_\ell, \texttt{R-BOCPD-UCRL2}, o_{c_\ell}, \left[c_\ell, c_{\ell+1}\right) \right) \nonumber \\
    & \hspace{-70pt} \leq 34 O \sqrt{A} \sum\limits_{\ell=0}^{K_T-1} D_\ell \sqrt{\left(c_{\ell+1}-c_{\ell}-d_{\ell} \right) \log\left(\frac{c_{\ell+1}-c_{\ell}-d_{\ell}}{\delta_\ell} \right)} + \sum\limits_{\ell=0}^{K_T-1} \kD_{\bDelta_{\ell+1},c_{\ell}+d_{\ell},c_{\ell+1}} \nonumber \\
    & \hspace{-70pt} = 34 D O \sqrt{A} \sum\limits_{\ell=0}^{K_T-1} \sqrt{\frac{1}{K_T} K_T \left(c_{\ell+1}-c_{\ell}-d_{\ell} \right) \log\left (\frac{K_T  (c_{\ell+1}-c_{\ell}-d_{\ell})}{\delta} \right)} + \sum\limits_{\ell=0}^{K_T-1} \kD_{\bDelta_{\ell+1},c_{\ell}+d_{\ell},c_{\ell+1}} \nonumber \\
    & \hspace{-70pt} \leq 34 D O \sqrt{A} \sum\limits_{\ell=0}^{K_T-1} \sqrt{\frac{T}{K_T} \log\left (\frac{T}{\delta} \right)} + \sum\limits_{\ell=0}^{K_T-1} \kD_{\bDelta_{\ell+1},c_{\ell}+d_{\ell},c_{\ell+1}} \nonumber \\
    & \hspace{-70pt} = 34 D O \sqrt{A T K_T \log\left (\frac{T}{\delta} \right)} + \sum\limits_{\ell=0}^{K_T-1} \kD_{\bDelta_{\ell+1},c_{\ell}+d_{\ell},c_{\ell+1}} \label{eq::ft-regret}
\end{align}

which holds with probability at least $1-\delta$. This completes the proof for Theorem \ref{main_theorem}.\\

\subsection{Asymptotic Regret Bound}
Now building up on the previous section and the asymptotic detection delay in Equation (\ref{eq::DetectionDelay}), we write the asymptotic regret bounds as follows 

\begin{align*}
    \mathfrak{R} \left(\mathbf{M}, \texttt{R-BOCPD-UCRL2}, o_{c_0}, T\right) & \leq 34 D O \sqrt{A T K_T \log\left (\frac{T}{\delta} \right)} + \sum\limits_{\ell=0}^{K_T-1} \kD_{\bDelta_{\ell+1},c_{\ell}+d_{\ell},c_{\ell+1}} \\
    & \leq 34 D O \sqrt{A T K_T \log\left (\frac{T}{\delta} \right)} + \sum\limits_{\ell=0}^{K_T-1} \lim_{c_{\ell + 1} - c_\ell - d_\ell \to \infty} \kD_{\bDelta_{\ell+1},c_{\ell}+d_{\ell},c_{\ell+1}} \\
    & = 34 D O \sqrt{A T K_T \log\left (\frac{T}{\delta} \right)} + \sum\limits_{\ell=0}^{K_T-1} \order{\frac{\log \frac{K_T}{\delta}}{\infdiv*{\btheta^{(\ell+1)}}{\btheta^{(\ell)}}}} \\
    & \leq 34 D O \sqrt{A T K_T \log\left (\frac{T}{\delta} \right)} + \order{\frac{K_T \log \frac{K_T}{\delta}}{\min\limits_\ell \: \infdiv*{\btheta^{(\ell+1)}}{\btheta^{(\ell)}}}}
\end{align*}

which again holds with probability at least $1-\delta$, assuming a false-alarm rate of $0$. This completes the derivation of Corollary \ref{main_cor}. 

\section{Experimental Setup \& Discussion}\label{sec:xps}
Given the generality of \texttt{R-BOCPD-UCRL2} with respect to variations in the state transition distributions and rewards, a suitable environment to benchmark its performance vis-à-vis state-of-the-art is a synthetic environment where abrupt changes occur to the state transition distributions and rewards at unknown time instances. First, the sizes of the state and action spaces is chosen randomly. Then, the state transition probabilities are sampled from a multinomial distribution over the set of states $\sO$ and the rewards are sampled randomly from $[0, 1]$, where we can control the variation in the generation process to be able to simulate both large changes to state-transition distributions and rewards and relatively small ones. Now, fixing a set of change-points, chosen with sufficient time difference in-between successive ones, the process is repeated after each change-point over a time horizon $T = 50000$. We consider 100 realizations of each state-action pair and are interested in the average cumulative rewards of that. 

\subsection{Choice of Hyperparameters}

Now, we list our hyperparameter choice for the sliding-window based algorithms in Section \ref{xps:main}. 
\begin{itemize}
    \item \textbf{Sliding-Window UCRL2} (\texttt{SWUCRL2},  \cite{SWUCRL}): The window size is chosen optimally as in \cite{SWUCRL}, $W^\star = \left(\frac{16.53}{K_T} T D O \sqrt{A \log{\frac{T}{\delta}}} \right)^\frac{2}{3}$. The diameter $D$ is estimated based on a hyperparameter search over a large range of suitable values for each combination of state space and action space sizes. The diameter maximizing the cumulative rewards was chosen. 
    \item \textbf{Sliding-Window UCRL2 with Confidence Widening} (\texttt{SWUCRL2-CW}, \cite{SWUCRL2CW}):  Again, the window size and widening factor are chosen optimally according to \cite{SWUCRL2CW}, $W^\star \dfq 3O^\frac{2}{3} A^\frac{1}{2} T^\frac{1}{2}/ (B_r + B_p + 1)^\frac{1}{2}$ and $\eta^\star \dfq \sqrt{(B_p + 1) W^\star / T}$ respectively. Here, $B_p$ and $B_r$ are computed beforehand in a total-variation sense. While, in a realistic RL setting, $T$ is also unknown beforehand, we still choose our choice of $T$ to initialize \texttt{SWUCRL2}. 
\end{itemize} 

\subsection{Discussion}\label{sec:xps_discussion}

Now, in addition to the rather general performance evaluation provided in Section \ref{xps:main}, we highlight in more detail how each algorithm operates as follows

\begin{itemize}
    \item \textbf{Sliding-Window UCRL2} (\texttt{SWUCRL2},  \cite{SWUCRL}): An essential parameter of the sliding-window approach is the diameter $D_\ell$ of each MDP $M_\ell$ defining the switching-MDP problem $\mathbf{M}$, which is used to quantify the difficulty of learning in the setting specified by $M_\ell$. Diameter $D_\ell$, as defined in Section \ref{sec: setting}, is a parameter that cannot be accessed directly from the MDP and yet is key for \texttt{SWUCRL2} to perform as claimed. Even while considering $D = \max\limits_{\ell} D_\ell$, \texttt{SWUCRL2} still relies on a hyperparameter search to estimate the overall optimal diameter $D$, which is quite restrictive in practice. We also highlight that \texttt{SWUCRL2} performs poorly for a suboptimal choice of $D$. Now, considering the optimal window-size choice $W^\star$ is of $\tilde{\mathcal{O}}(A^{\frac{1}{3}} O^{\frac{2}{3}} D^{\frac{2}{3}} T^{\frac{2}{3}} {K_T}^{-\frac{1}{2}})$, \texttt{SWUCRL2} requires to keep track of a significant number of observations even for rather small MDPs. 
    \item \textbf{Sliding-Window UCRL2 with Confidence Widening} (\texttt{SWUCRL2-CW}, \cite{SWUCRL2CW}): While not relying on an agnostically chosen parameter as for \texttt{SWUCRL2}, (\texttt{SWUCRL2-CW} still relies on the knowledge of predefined variation budgets for the state-transition distributions and rewards, which are unknown in a realistic setting. Its variant Bandit-over-Reinforcement Learning (\texttt{BORL}), which operates without assuming the knowledge of $B_r$ and $B_p$, performs than \texttt{SWUCRL2-CW} in practice. 
    \item \textbf{Restarted UCRL2} (\texttt{Restarted-UCRL2}, \cite{UCRL2}): While comparing favorably to sliding-window approaches, it requires a very large number of restarts $T^{\frac{1}{3}} {K_T}^{\frac{2}{3}}$, which is quite prohibitive for large-sizes problems. In addition, since restarting frequency decreases polynomially with time (for a fixed number of changes),  performance will inevitably degrade for considerably long time horizons with changes occuring frequently at the later stages of learning. 
    \item \textbf{Vanilla UCRL2} (\texttt{UCRL2}, \cite{UCRL2}): Here rather considered as a baseline. 
\end{itemize}

\subsubsection{Extension to realistic environments}
We also highlight our interest in realistic MDP settings such as \textit{RiverSwim} (\cite{Rswim}), \textit{MachineReplacement} and \textit{GridWorld} among others. Here, the difficulty of simulating realistic changes to the environment is that we don't have access to the underlying state-transition distributions and rewards. While it is possible to alter some of the environment parameters that are typically chosen at random, it is unclear to us at the moment of writing this manuscript how to control the process of changing these variables and relate it to the rather latent changes in the underlying state-transition distributions and rewards. 



\end{document}

%% file: math_commands2.tex
\usepackage{amsmath,amsfonts,bm}
\usepackage{amsbsy}

\usepackage{calrsfs}
\DeclareMathAlphabet{\pazocal}{OMS}{zplm}{m}{n}

\newcommand{\pA}{\pazocal{A}}
\newcommand{\pB}{\pazocal{B}}
\newcommand{\pC}{\pazocal{C}}

\newcommand{\pG}{\pazocal{G}}

\newcommand{\pM}{\pazocal{M}}
\newcommand{\pW}{\pazocal{W}}

\DeclareMathAlphabet{\mathsfit}{\encodingdefault}{\sfdefault}{m}{sl}
\SetMathAlphabet{\mathsfit}{bold}{\encodingdefault}{\sfdefault}{bx}{n}

\def\gA{{\mathcal{A}}}

\def\sA{{\mathsf{A}}}

\def\sO{{\mathsf{O}}}



%% file: nonStationaryMDP.tikz
  
\tikzset {_05agoytyn/.code = {\pgfsetadditionalshadetransform{ \pgftransformshift{\pgfpoint{0 bp } { 0 bp }  }  \pgftransformrotate{0 }  \pgftransformscale{2 }  }}}
\pgfdeclarehorizontalshading{_s8191ilbc}{150bp}{rgb(0bp)=(0.95,0.91,0.4);
rgb(37.5bp)=(0.95,0.91,0.4);
rgb(62.5bp)=(1,0.71,0.27);
rgb(100bp)=(1,0.71,0.27)}

  
\tikzset {_qli48u6zh/.code = {\pgfsetadditionalshadetransform{ \pgftransformshift{\pgfpoint{0 bp } { 0 bp }  }  \pgftransformrotate{0 }  \pgftransformscale{2 }  }}}
\pgfdeclarehorizontalshading{_zhjb0l71p}{150bp}{rgb(0bp)=(0.6,0.85,1);
rgb(37.5bp)=(0.6,0.85,1);
rgb(62.5bp)=(0,0.5,0.5);
rgb(100bp)=(0,0.5,0.5)}

  
\tikzset {_3ffwztb3q/.code = {\pgfsetadditionalshadetransform{ \pgftransformshift{\pgfpoint{0 bp } { 0 bp }  }  \pgftransformrotate{0 }  \pgftransformscale{2 }  }}}
\pgfdeclarehorizontalshading{_19jchb8kn}{150bp}{rgb(0bp)=(1,0,0);
rgb(37.5bp)=(1,0,0);
rgb(50bp)=(1,1,0);
rgb(62.5bp)=(1,0,0);
rgb(100bp)=(1,0,0)}

  
\tikzset {_517yv3s68/.code = {\pgfsetadditionalshadetransform{ \pgftransformshift{\pgfpoint{0 bp } { 0 bp }  }  \pgftransformrotate{0 }  \pgftransformscale{2 }  }}}
\pgfdeclarehorizontalshading{_c02m1t427}{150bp}{rgb(0bp)=(0.8,0.38,0.7);
rgb(37.5bp)=(0.8,0.38,0.7);
rgb(50bp)=(0.68,0.07,0.51);
rgb(62.5bp)=(0.87,0.28,0.67);
rgb(100bp)=(0.87,0.28,0.67)}
\tikzset{every picture/.style={line width=0.75pt}} 

\begin{tikzpicture}[x=0.75pt,y=0.75pt,yscale=-1,xscale=1]

\path  [shading=_s8191ilbc,_05agoytyn] (47,131.4) .. controls (47,127.31) and (50.31,124) .. (54.4,124) -- (201.6,124) .. controls (205.69,124) and (209,127.31) .. (209,131.4) -- (209,153.6) .. controls (209,157.69) and (205.69,161) .. (201.6,161) -- (54.4,161) .. controls (50.31,161) and (47,157.69) .. (47,153.6) -- cycle ; 
 \draw  [line width=1.5]  (47,131.4) .. controls (47,127.31) and (50.31,124) .. (54.4,124) -- (201.6,124) .. controls (205.69,124) and (209,127.31) .. (209,131.4) -- (209,153.6) .. controls (209,157.69) and (205.69,161) .. (201.6,161) -- (54.4,161) .. controls (50.31,161) and (47,157.69) .. (47,153.6) -- cycle ; 

\draw [color={rgb, 255:red, 74; green, 144; blue, 226 }  ,draw opacity=1 ][fill={rgb, 255:red, 74; green, 144; blue, 226 }  ,fill opacity=1 ][line width=2.25]    (53,183) -- (663,178.03) ;
\draw [shift={(667,178)}, rotate = 179.53] [color={rgb, 255:red, 74; green, 144; blue, 226 }  ,draw opacity=1 ][line width=2.25]    (34.98,-10.53) .. controls (22.24,-4.47) and (10.58,-0.96) .. (0,0) .. controls (10.58,0.96) and (22.24,4.47) .. (34.98,10.53)   ;
\path  [shading=_zhjb0l71p,_qli48u6zh] (224,131.4) .. controls (224,127.31) and (227.31,124) .. (231.4,124) -- (344.6,124) .. controls (348.69,124) and (352,127.31) .. (352,131.4) -- (352,153.6) .. controls (352,157.69) and (348.69,161) .. (344.6,161) -- (231.4,161) .. controls (227.31,161) and (224,157.69) .. (224,153.6) -- cycle ; 
 \draw  [line width=1.5]  (224,131.4) .. controls (224,127.31) and (227.31,124) .. (231.4,124) -- (344.6,124) .. controls (348.69,124) and (352,127.31) .. (352,131.4) -- (352,153.6) .. controls (352,157.69) and (348.69,161) .. (344.6,161) -- (231.4,161) .. controls (227.31,161) and (224,157.69) .. (224,153.6) -- cycle ; 

\path  [shading=_19jchb8kn,_3ffwztb3q] (363,132.4) .. controls (363,128.31) and (366.31,125) .. (370.4,125) -- (474.6,125) .. controls (478.69,125) and (482,128.31) .. (482,132.4) -- (482,154.6) .. controls (482,158.69) and (478.69,162) .. (474.6,162) -- (370.4,162) .. controls (366.31,162) and (363,158.69) .. (363,154.6) -- cycle ; 
 \draw  [line width=1.5]  (363,132.4) .. controls (363,128.31) and (366.31,125) .. (370.4,125) -- (474.6,125) .. controls (478.69,125) and (482,128.31) .. (482,132.4) -- (482,154.6) .. controls (482,158.69) and (478.69,162) .. (474.6,162) -- (370.4,162) .. controls (366.31,162) and (363,158.69) .. (363,154.6) -- cycle ; 

\path  [shading=_c02m1t427,_517yv3s68] (495,131.4) .. controls (495,127.31) and (498.31,124) .. (502.4,124) -- (649.6,124) .. controls (653.69,124) and (657,127.31) .. (657,131.4) -- (657,153.6) .. controls (657,157.69) and (653.69,161) .. (649.6,161) -- (502.4,161) .. controls (498.31,161) and (495,157.69) .. (495,153.6) -- cycle ; 
 \draw  [line width=1.5]  (495,131.4) .. controls (495,127.31) and (498.31,124) .. (502.4,124) -- (649.6,124) .. controls (653.69,124) and (657,127.31) .. (657,131.4) -- (657,153.6) .. controls (657,157.69) and (653.69,161) .. (649.6,161) -- (502.4,161) .. controls (498.31,161) and (495,157.69) .. (495,153.6) -- cycle ; 

\draw [color={rgb, 255:red, 208; green, 2; blue, 27 }  ,draw opacity=1 ][line width=1.5]    (220,172.5) -- (220,195.5) ;
\draw [color={rgb, 255:red, 208; green, 2; blue, 27 }  ,draw opacity=1 ][line width=1.5]    (490,168.5) -- (490,191.5) ;
\draw [color={rgb, 255:red, 208; green, 2; blue, 27 }  ,draw opacity=1 ][line width=1.5]    (360,169) -- (360,192) ;
\draw [color={rgb, 255:red, 208; green, 2; blue, 27 }  ,draw opacity=1 ][line width=1.5]    (200,248) .. controls (239.2,218.6) and (182.36,225.22) .. (217.72,197.26) ;
\draw [shift={(220,195.5)}, rotate = 143.13] [color={rgb, 255:red, 208; green, 2; blue, 27 }  ,draw opacity=1 ][line width=1.5]    (14.21,-4.28) .. controls (9.04,-1.82) and (4.3,-0.39) .. (0,0) .. controls (4.3,0.39) and (9.04,1.82) .. (14.21,4.28)   ;
\draw [color={rgb, 255:red, 208; green, 2; blue, 27 }  ,draw opacity=1 ][line width=1.5]    (200,248) .. controls (239.8,218.65) and (347.91,301.66) .. (359.83,193.64) ;
\draw [shift={(360,192)}, rotate = 95.68] [color={rgb, 255:red, 208; green, 2; blue, 27 }  ,draw opacity=1 ][line width=1.5]    (14.21,-4.28) .. controls (9.04,-1.82) and (4.3,-0.39) .. (0,0) .. controls (4.3,0.39) and (9.04,1.82) .. (14.21,4.28)   ;
\draw [color={rgb, 255:red, 208; green, 2; blue, 27 }  ,draw opacity=1 ][line width=1.5]    (200,248) .. controls (239.8,218.15) and (412.26,326.41) .. (488.85,193.52) ;
\draw [shift={(490,191.5)}, rotate = 119.2] [color={rgb, 255:red, 208; green, 2; blue, 27 }  ,draw opacity=1 ][line width=1.5]    (14.21,-4.28) .. controls (9.04,-1.82) and (4.3,-0.39) .. (0,0) .. controls (4.3,0.39) and (9.04,1.82) .. (14.21,4.28)   ;
\draw [color={rgb, 255:red, 208; green, 2; blue, 27 }  ,draw opacity=1 ][line width=1.5]    (53,171.5) -- (53,194.5) ;

\draw (72,133) node [anchor=north west][inner sep=0.75pt]   [align=left] {\textbf{{\large Model $1$}} ($M_0$)};
\draw (237,133) node [anchor=north west][inner sep=0.75pt]   [align=left] {\textbf{{\large Model $2$}} ($M_1$)};
\draw (371,134) node [anchor=north west][inner sep=0.75pt]   [align=left] {\textbf{{\large Model $3$}} ($M_2$)};
\draw (520,133) node [anchor=north west][inner sep=0.75pt]   [align=left] {\textbf{{\large Model $4$}} ($M_3$)};
\draw (382,213) node [anchor=north west][inner sep=0.75pt]   [align=left] {\textbf{{\large \textcolor[rgb]{0.29,0.56,0.89}{TIME}}}};
\draw (134,258) node [anchor=north west][inner sep=0.75pt]   [align=left] {\textbf{\textcolor[rgb]{0.82,0.01,0.11}{Unknown change-point}}};
\draw (222,198.9) node [anchor=north west][inner sep=0.75pt]    {$c_{1}$};
\draw (329,193.9) node [anchor=north west][inner sep=0.75pt]    {$c_{2}$};
\draw (492,194.9) node [anchor=north west][inner sep=0.75pt]    {$c_{3}$};
\draw (50,200.9) node [anchor=north west][inner sep=0.75pt]    {$c_{0}$};

\end{tikzpicture}